\def\year{2018}\relax
\documentclass[letterpaper]{article} 
\usepackage{aaai18}  
\usepackage{times}  
\usepackage{helvet}  
\usepackage{courier}  
\usepackage{url}  
\usepackage{graphicx}  
\usepackage{epsfig}
\usepackage{graphicx}
\usepackage{amsmath}
\usepackage{amssymb}
\usepackage{setspace}
\usepackage{subcaption}
\usepackage{grffile}
\usepackage{verbatim}
\usepackage{booktabs}
\usepackage{placeins}
\usepackage[skip=0pt]{caption}
\usepackage{array}
\usepackage[usenames,dvipsnames]{color}
\usepackage{setspace}
\usepackage[inline]{enumitem}

\usepackage[pagebackref=true,breaklinks=true,linkcolor={},bookmarks=false]{hyperref}

\newcommand{\myparagraph}[1]{
\vspace{0.1cm}\noindent
\textbf{#1}
\hspace{0.1cm}
}

\newcolumntype{C}[1]{>{\centering\let\newline\\\arraybackslash\hspace{0pt}}m{#1}}

\usepackage{color}
\begin{document}
%
\title{Long-Term Image Boundary Prediction}
\author{Apratim Bhattacharyya, Mateusz Malinowski, Bernt Schiele, Mario Fritz \\ 
Max Planck Institute for Informatics \\
Saarland Informatics Campus, Saarbr\"{u}cken, Germany \\
\texttt{\{abhattac, mmalinow, schiele, mfritz\}@mpi-inf.mpg.de}  }
\maketitle


\begin{abstract}
Boundary estimation in images and videos has been a very active topic of research, and organizing visual information into boundaries and segments is believed to be a corner stone of visual perception. While prior work has focused on estimating boundaries for observed frames, our work aims at predicting boundaries of future unobserved frames. This requires our model to learn about the fate of boundaries and corresponding motion patterns -- including a notion of ``intuitive physics''. We experiment on natural video sequences along with synthetic sequences with deterministic physics-based and agent-based motions. While not being our primary goal, we also show that fusion of RGB and boundary prediction leads to improved  RGB predictions.
\end{abstract}


\section{Introduction}
Humans possess the skill to imagine future states of observed scenes in  diverse scenarios. This supports various different tasks ranging from planning to object manipulation, e.g. a goalkeeper jumping to intercept the ball or reaching out for a handshake. Humans can readily perform such complex and versatile tasks because they can anticipate motions including an intuitive understanding of physical laws from the early age \cite{baillargeon1994infants,baillargeon2004infants}. 

\begin{figure}[t]
\centering
\begin{tabular}{C{3.4cm}C{3.4cm} }
\includegraphics[width=0.20\textwidth, height=0.08\textheight]{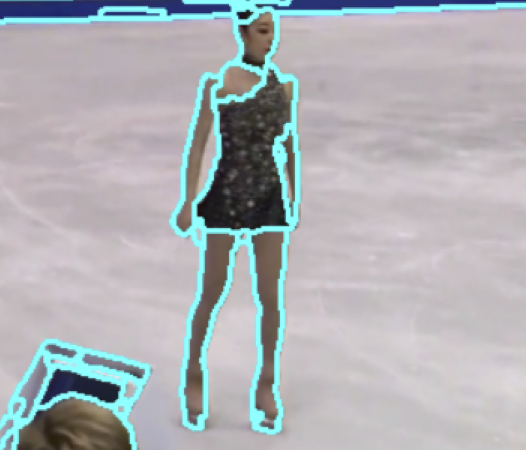} &
\includegraphics[width=0.20\textwidth, height=0.08\textheight]{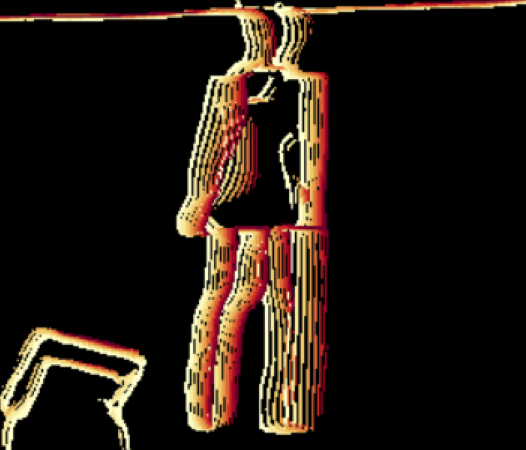}\\
\addlinespace[-2ex]
\includegraphics[width=0.20\textwidth, height=0.08\textheight]{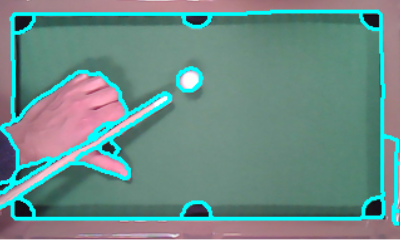} &
\includegraphics[width=0.20\textwidth, height=0.08\textheight]{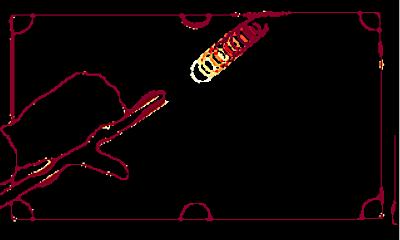}\\
\textbf{Last Observation: $t$} & \textbf{Prediction} \\
\end{tabular}
\caption{Predicted future boundary images, from $t$ + 1 (Yellow) to $t$ + 8 (Row 1), $t$ + 18 (Row 2) (Red), superimposed.}\label{fig:teaser}
\end{figure}

In this work, we propose the task of predicting future scene boundaries. Scene boundaries capture the important structure and extents of objects. Moreover, they can be accurately estimated \cite{khoreva2016improved}. Prediction of future scene boundaries requires understanding of object dynamics and motion patterns including an intuitive understanding of physical laws or ``intuitive physics''. In this work, we focus on two particular scenarios involving motion and local interactions. The first one, which we call physics-based motion, can fully be described by the laws of physics, e.g. dynamics of  billiard balls. The second one, which we call agent-based motion, also involves understanding of intentions, e.g. dynamics of an ice-skater. Therefore, our methods have to deal with diverse situations, work on raw pixels, and should be capable of long-term predictions. Figure \ref{fig:teaser} shows example results of our method that accurately predicts future scene boundaries.

Recently, full future frame predication of observed scenes has been studied \cite{mathieu2015deep,liu2017video}. But up to now, only very short range predictions of few frames have been shown, where blurriness/distortion artifacts occur in the predicted future frames -- losing/incorrectly propagating high-frequency information. This high frequency information is crucial for meaningful predictions about the future, e.g. on a billiard table the location of a ball and table boundaries are necessary to infer the future state of the table. Boundaries capture this crucial high frequency information and are also known to reveal important structures of the visual scene \cite{wertheimer1923laws,amfm_pami2011,galasso2013unified}. Therefore, we argue that the task of future boundary prediction is a more suitable benchmark for understanding and predicting physics or agent-based motion.

Our main contributions are as follows, 
\begin{enumerate*}
    \item We propose the novel task of future boundary prediction.
    \item We propose the first method that predicts future boundaries based only on the raw pixels.
    \item We evaluate our model on two scenarios involving physics-based (synthetic and real billiard sequences) and agent-based motion (VSB100, \cite{galasso2013unified}).
    \item Under the physics-based scenario, the method shows for the first time long-term predictions.
    \item Under the agent-based scenario on VSB100 and UCF101, we show that the predicted boundaries can be used in a fusion scheme that improves RGB video prediction in the longer-term.
\end{enumerate*}


\begin{figure*}[t]
    \centering
    \includegraphics[width=0.8\linewidth]{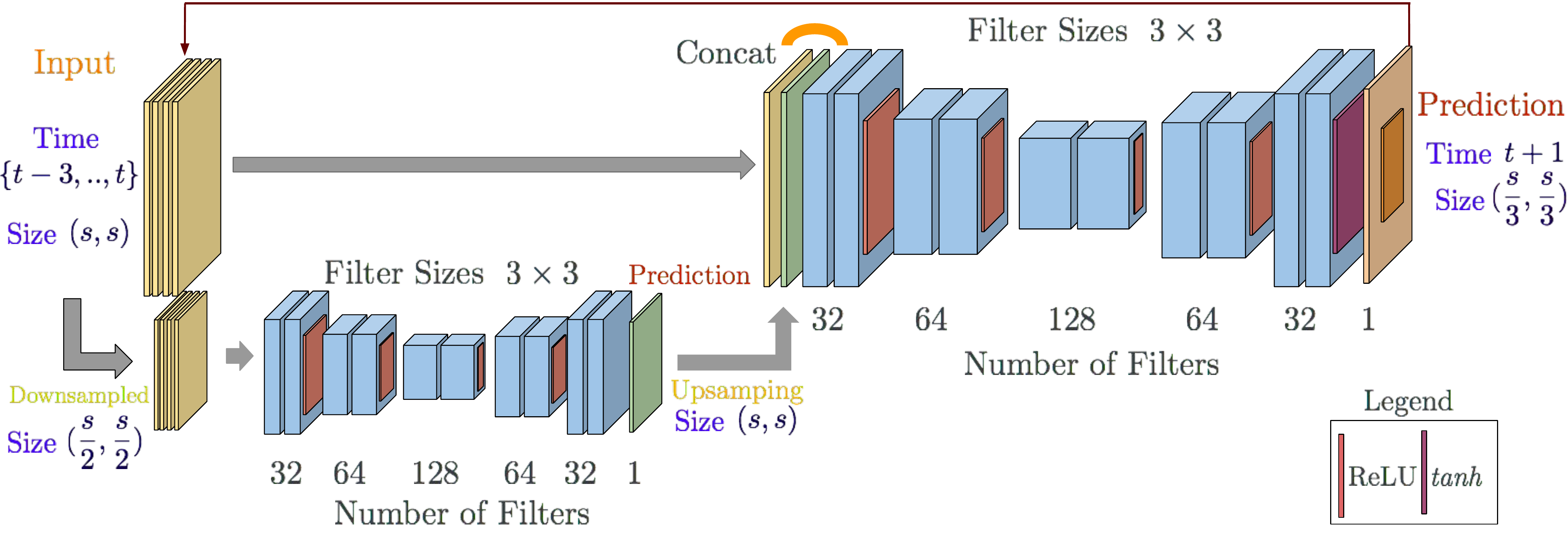}    
    \caption{Convolutional Multi-scale with Context architecture (only 2 out of 4 scales illustrated).}
    \label{fig:modelarch}
\end{figure*}

\section{Related work} \label{sec:relatedwork}
\myparagraph{RGB frame prediction.} This problem has  recently received a lot of attention. However, the predicted frames have blurriness problems. \cite{ranzato2014video} sought to remedy this problem by discretizing the input through k-means atoms and predicting on this vocabulary instead. The work of \cite{mathieu2015deep} proposes using adversarial loss, which leads to improved results over \cite{ranzato2014video}. \cite{liang2017dual,liu2017video,patraucean2015spatio} shows some further improvement through the use of optical flow information. However, these approaches produce sharper short term predictions but still suffer from blurriness problems starting as soon as 3 frames into the future. \cite{kalchbrenner2016video} focus on moving MNIST digits and like \cite{finn2016unsupervised} on action conditioned video prediction.  \cite{villegas2017learning} proposes a hierarchical approach for making long-term frame predictions, by first estimating the high-level structure in the input frames and predicting how that structure evolves in the future. They show promising results on videos where pose is an easily identifiable and appropriate high level structure to exploit. However, such high-level structures are video domain dependent. Other works  \cite{sutskever2009recurrent,michalski2014modeling} focus on deterministic bouncing ball sequences, but their dataset is limited in size and resolution and generalization with respect to the number of balls and their velocities is not considered.

\myparagraph{Intuitive physics.} Developing an intuitive understanding of physics from raw visual input has been explored recently. \cite{fragkiadaki2015learning} predict future states of balls moving on a billiard table and \cite{lerer2016learning,li2016fall} predict the stability of towers made out of blocks.  However, both \cite{fragkiadaki2015learning} and \cite{lerer2016learning} have an ``object notion'', meaning that the architecture knows a priori the location or type of the objects that  it is supposed to infer. Although some recent approaches such as \cite{battaglia2016interaction,watters2017visual} are capable of long-term predictions, they are modeling either state-to-state or images-to-state transitions. Moreover, in the latter case, the input is visually simplified, and the focus is only on deterministic motions. In contrast to this body of work, we focus on  more diverse scenarios and are agnostic to the underlying objects and causes of change. 

\myparagraph{Video segmentation.} Video segmentation as the task of finding consistent spatio-temporal boundaries in a video volume has received significant attention over the last years \cite{galasso2014spectral,ochs2014segmentation,galasso2013unified,chang2013video}, as it provides an initial analysis and abstraction for further processing. In contrast, our approach aims at predicting these boundaries into the future without any video observed for  future frames. 


\section{Model} \label{sec:models}

We present a model that observes a sequence of boundary images, where each pixel encodes the confidence of occurrence of an image boundary at that location and then predicts the boundary image(s) at the next time-step(s). 
An overview of our Convolutional Multi-Scale Context (CMSC) model is shown in Figure \ref{fig:modelarch}. 

We approach long term prediction by recursion, due to the advantage of efficiency. However, errors are potentially propagated and accumulated over time. To mitigate such effects, we need our model to be accurate and to consolidate information over time. To maximize accuracy, our model has been designed through analysis of prior work on the related task of frame prediction. Furthermore, our model has many novel aspects which are key to long term prediction.

In order to generalize across diverse sequences while maintaining a tractable number of parameters, a patch based approach is adopted. Therefore, our model observes and predict on patches rather than the complete input image. Alternatively, this can be seen as multiple replicas (``patch predictors'') of our model predicting on patches of the input sequence. We now describe our model through an analysis of its various components.

\subsection{Fully Convolutional.} Our CMSC model consists of only convolutional layers. The input boundary image sequence is concatenated as channels and is read by the first convolutional layer. Convolutional layers can extract high quality location invariant features. In particular, they can extract information about the orientation and direction of motion of boundaries. Neurons at upper convolutional layers have larger receptive fields and can aggregate information. In fact, as shown by the work  \cite{jain2007supervised}, the output layer should have a wide receptive field to preserve long range spatial and temporal dependencies and learn about interaction among boundaries in a spatio-temporal context. We therefore use several convolutional layers in our CMSC model. We also introduce pooling in between convolutional layers. Pooling further helps in the aggregation of information and increases receptive fields. However, excessive pooling (or tight bottlenecks with fully connected layers) have been shown to be successful in classification tasks, but also have shown by \cite{ranzato2014video} to induce image degradations for synthesis tasks. Therefore, it is crucial to use moderate pooling. Finally, we use up-sampling layers after pooling to maintain resolution. 

\subsection{Multiple Scale Prediction.} 
Multi-Scale model architectures akin to a Laplacian pyramid have shown to be advantageous for generating natural images \cite{denton2015deep} and predicting future RGB frames \cite{mathieu2015deep}. Such model architectures contain multiple levels which observes the input boundary image(s) at increasing (coarse to fine) scales. Down-sampling a boundary image has the effect of smoothing and discarding details of a boundary image. It is easier to predict future boundary images at a coarser resolutions. Therefore, our CMSC model uses multiple scales (or levels). The input $\text{I}(\text{L}_{2k})$ to a certain level ($\text{L}_{2k}$) is the input boundary image sequence scaled  to the current level $\text{X}_{2k}$ and the boundary image $\text{O}$ predicted by the previous coarser level ($\text{L}_{k}$). The boundary image predicted by the coarser level is upsampled $\hat{\text{O}}$ to the scale at the current level. We have,
\begin{align*}
    \text{I}(\text{L}_{2k}) &=  \left\{ \text{X}_{2k}, \hat{\text{O}}(\text{L}_{k}) \right\}
\end{align*} 
The coarse predicted boundary images $\hat{\text{O}}(\text{L}_{k})$ act a guide for the next higher level of the model. We use four levels, with scales increasing by a factor of two. 

\myparagraph{Details of each Level in our Model.}  Each level of the model consists of five sets of two convolutional layers. There are 32, 64, 128, 64 and 32 filters respectively in each set, of a constant size 3$\times$3. Multiple convolutional layers at each scale leads to large receptive fields at the output layer. We introduce moderate 2$\times$2 pooling layer after the first two sets of convolutional layers, leading to futher aggregation of information and increased receptive fields. We double the number of convolutional filters after pooling to aid feature extraction. We upsample the convolutional maps after the third set to maintain resolution. We use \emph{ReLU} non-linearities between every layer expect the last. We use the \emph{tanh} non-linearity at the end to ensure output in the range [0,1]. (Additional details in the Appendix)

For accurate long term prediction, it is crucial to ensure global consistency through communication between the patch predictors. Consider a video of a moving ball. The trajectory of a ball might intersect with multiple patches. To correctly predict the motion far into the future, replicas of the model predicting on neighboring patches need to be consistent especially during transition of the ball between patches. Therefore, we describe next the final component of our CMSC model, the context, which ensures global consistency.

\subsection{Context.} 
Our CMSC model observes a central patch along with the directly neighbouring 8 patches. This neighbourhood is called the context. However, the model only predicts on the central patch. While predicting recursively, the model observes its previous output along with the the output of the neighboring patch predictors. This enables the learning of spatially consistent predictions while keeping the same number of parameters.

\begin{figure}[h]
    \centering
    \includegraphics[width = 0.28\textwidth]{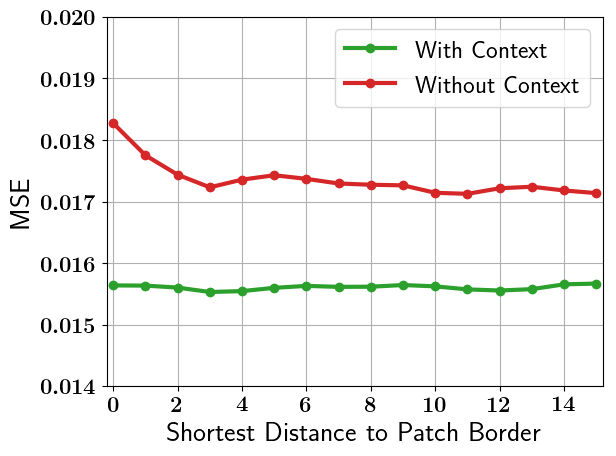}
    \caption{Our model without context has higher error near the patch boundary (red) vs. with context (green).}
    \label{fig:errordist}
\end{figure}

The addition of a context has the added advantage that the output layer neurons now have receptive fields that are uniform in size. Without context, the neurons at the boundary of the (2D) output layer have a smaller receptive field compared to the neurons at the center. This leads to a non-uniform (training and test) error distribution at the output layer neurons. In Figure \ref{fig:errordist} we plot the average error at the output layer neurons of our CMSC model at increasing distance from the patch border, with and without context. Error increases consistently from patch center (right) to the patch border (left) without a context. Note that, the model of \cite{mathieu2015deep} is also multi-scale and fully convolutional like CMSC, but it does not have pooling or context.

Next, we evaluate our CMSC model and the effectiveness of its various components.


\section{Experiments}
We evaluate our CMSC model on natural video sequences involving agent-based motion and billiard sequences with only physics-based motion. We compare with various baselines and perform ablation studies to confirm design choices. We convert each video into 32$\times$32 pixel patches. The CMSC model observes a central patch and eight neighbouring patches resulting in a context of size 96$\times$96 pixels. 

\begin{figure*}[!t]
    \centering
    \begin{subfigure}{0.34\textwidth}
    \includegraphics[width = \textwidth]{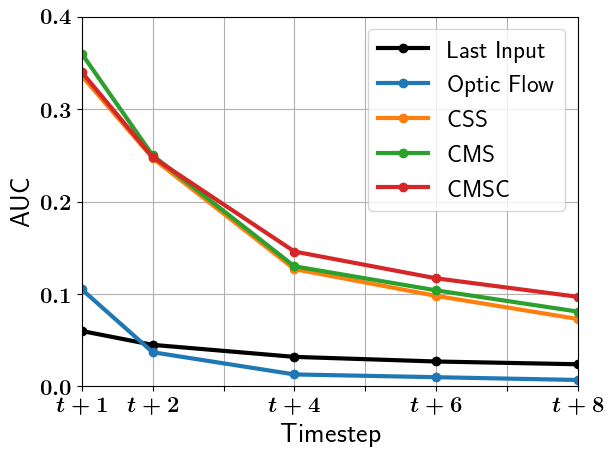}
    \caption{Area under the curve.}
    \label{fig:aucvsb100}
    \end{subfigure}
    \begin{subfigure}{0.34\textwidth}
    \includegraphics[width =\textwidth]{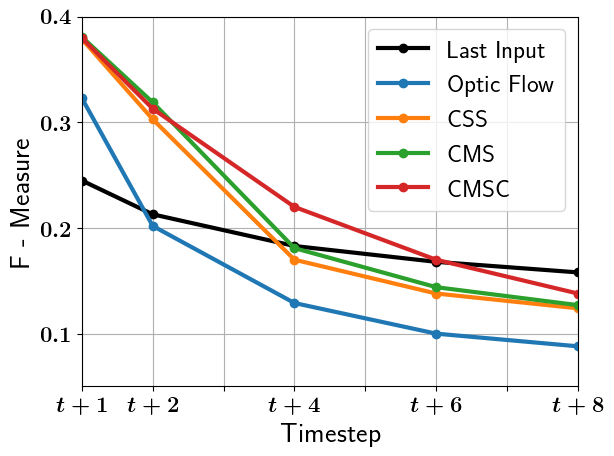}
    \caption{Best F-measure.}
    \label{fig:bestfvsb100}
    \end{subfigure}
    \begin{subfigure}{0.3\textwidth}
    \centering
        \begin{subfigure}{\linewidth}
        \centering
        \includegraphics[width = 0.65\textwidth]{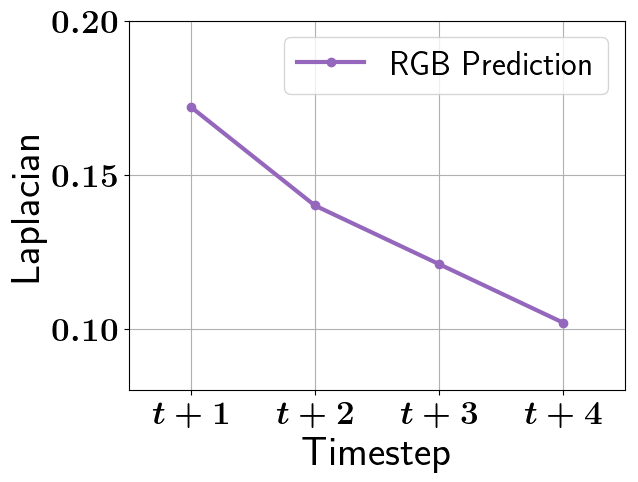}
        \caption{Laplacian measure.}
        \label{fig:shapness}
        \end{subfigure}
        \begin{subfigure}{\linewidth}
        \centering
        \includegraphics[width = 0.65\textwidth]{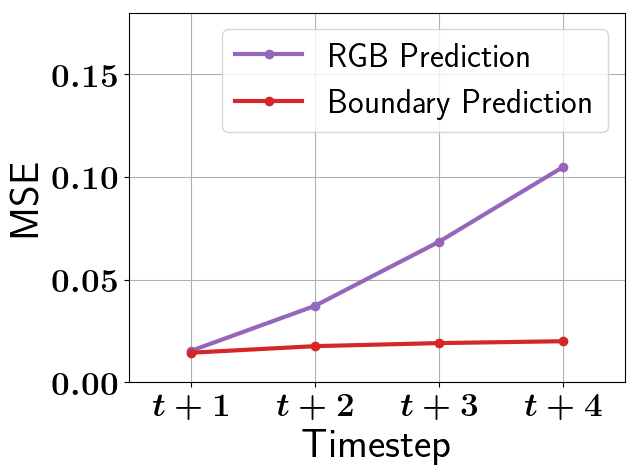}
        \caption{Mean squared error.}
        \label{fig:mse}
        \end{subfigure}
    \end{subfigure}
    \caption{Left and center: Evaluation of boundary prediction on VSB100. Right: RGB versus boundary prediction.}
\end{figure*}

\begin{figure*}[t]
  
  \centering
  \begin{tabular}{ cccc } 
     
    \includegraphics[width=0.21\textwidth]{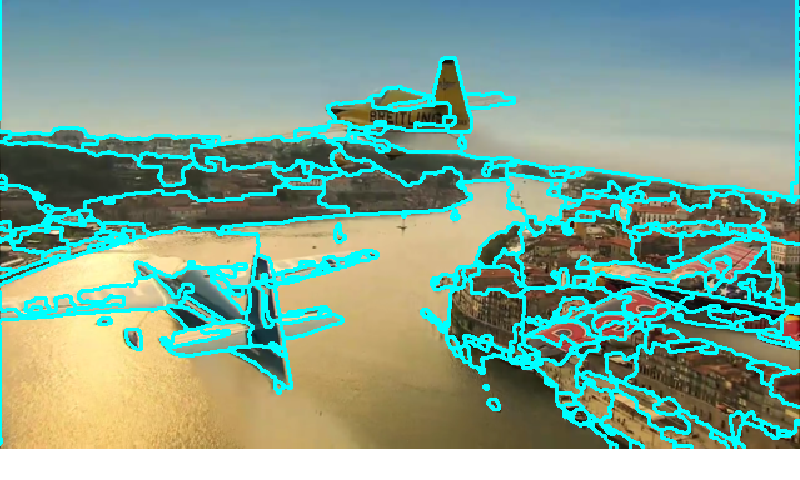} &
    \includegraphics[width=0.21\textwidth]{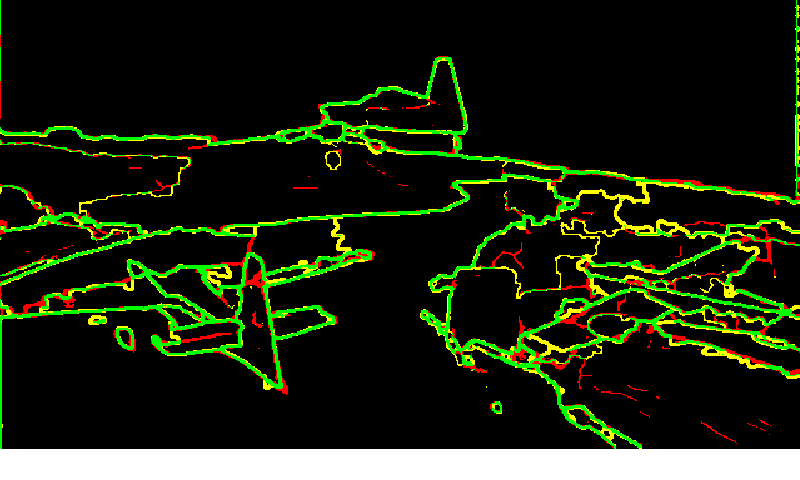} &
    \includegraphics[width=0.21\textwidth]{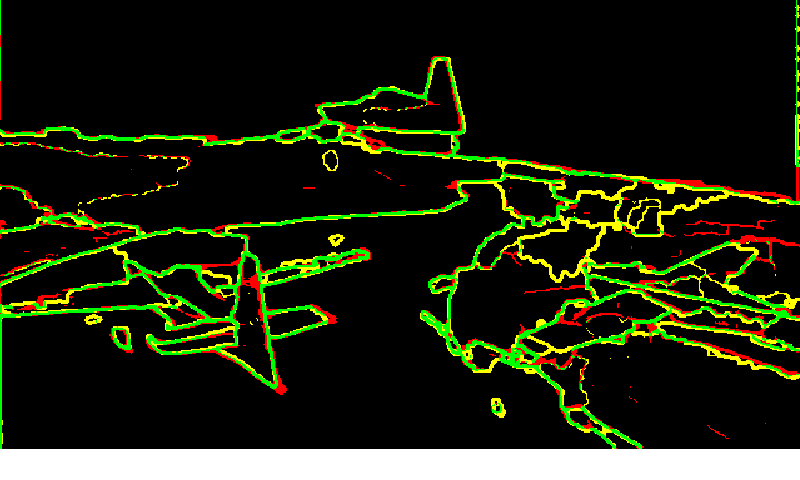} &
    \includegraphics[width=0.21\textwidth]{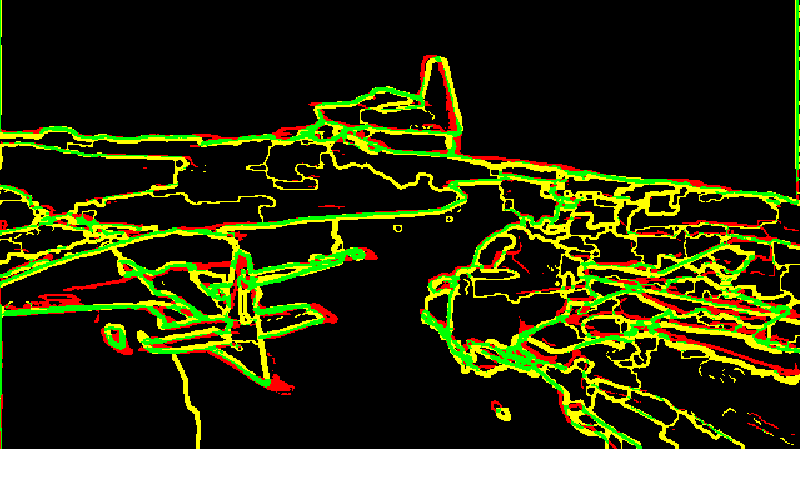} \\
    \includegraphics[width=0.21\textwidth]{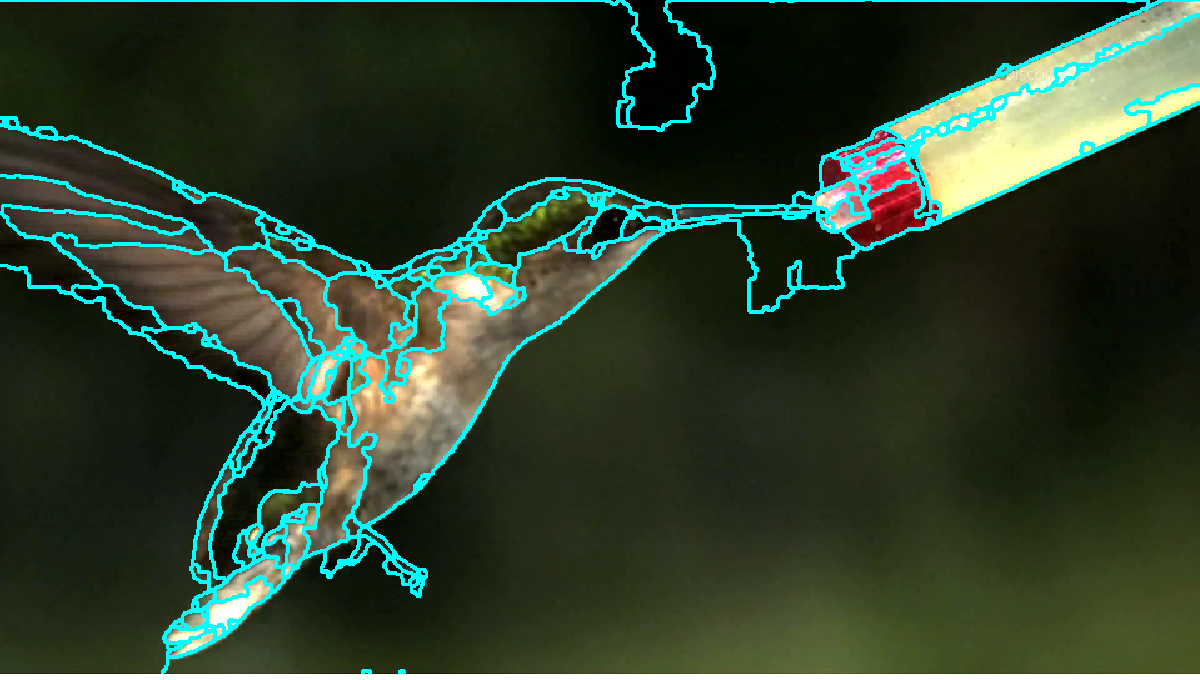}&
    \includegraphics[width=0.21\textwidth]{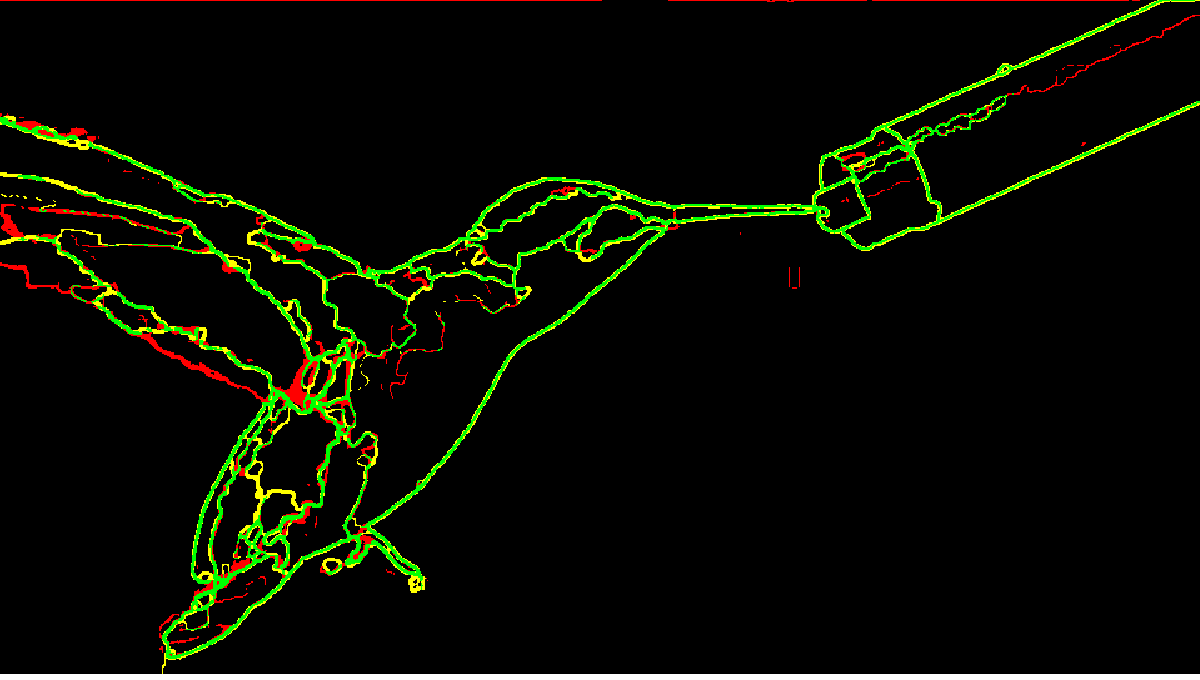}&
    \includegraphics[width=0.21\textwidth]{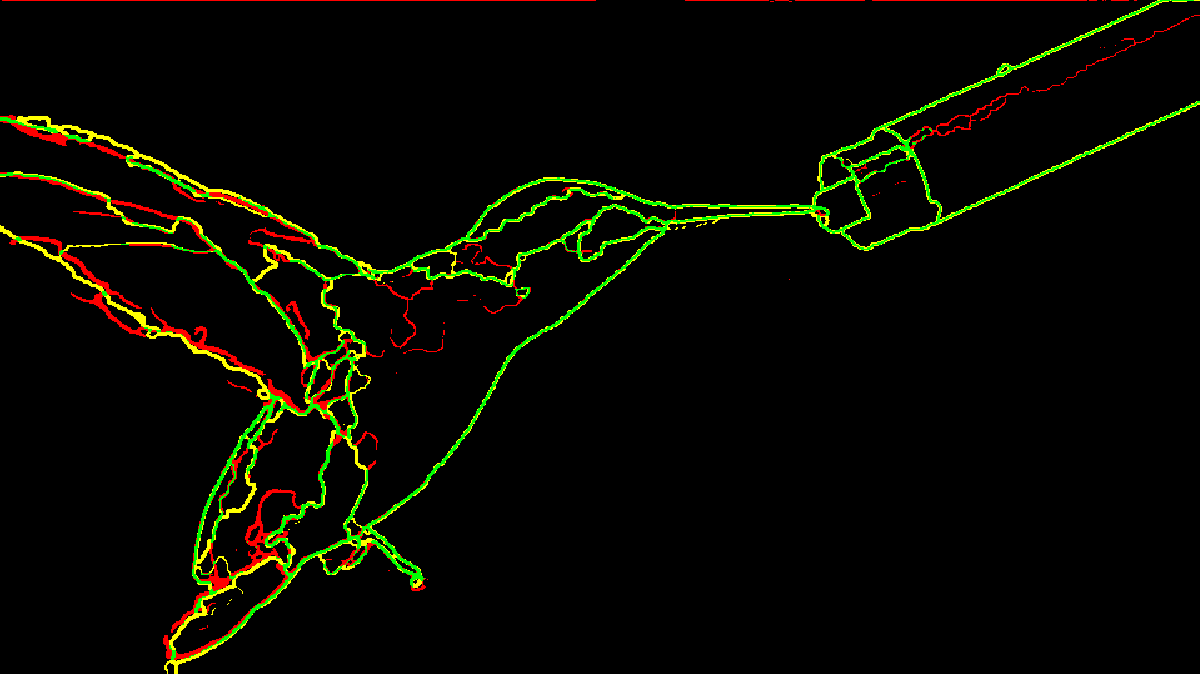}&
    \includegraphics[width=0.21\textwidth]{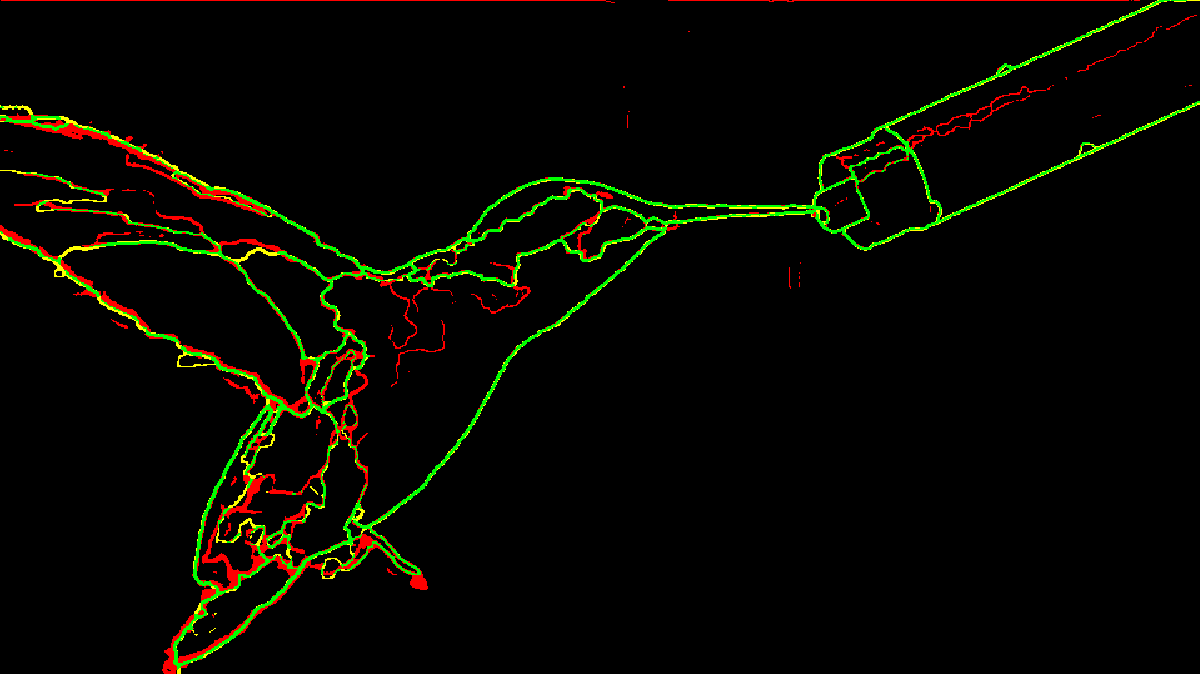}\\
    
    Last Observation: $t$ & Prediction: $t$ + 1 & Prediction: $t$ + 2 &  Prediction: $t$ + 4 \\
    
    \end{tabular}
  \caption{Rows top to bottom: Prediction on \emph{airplane} and \emph{hummingbird} sequences from VSB100. Correct boundaries predictions are encoded in green. Missed boundaries are encoded in yellow. Wrong boundaries are encoded in red.}
  \label{fig:cframes}
\end{figure*}

\myparagraph{Training Loss.} We use L2 loss (mean square error) during training, which we optimize using the ADAM optimizer. 

\myparagraph{Evaluation Metric.} As we want sharp and accurate boundaries, we use the established boundary precision recall (BPR) evaluation metric from the video segmentation literature \cite{galasso2013unified}. This metric is defined for a set $P$ of predicted boundary images and $G$ of corresponding ground truth boundary images as,
\begin{align*}
    P &= \frac{ \sum_{B_{p} \in P, B_{g} \in G} \mid B_{p} \cap B_{g} \mid }{\sum_{B_{p} \in P} \mid B_{p} \mid}\\
    R &= \frac{ \sum_{B_{p} \in P, B_{g} \in G} \mid B_{p} \cap B_{g} \mid }{\sum_{B_{g} \in G} \mid B_{g} \mid}\\
    F &= \frac{2 PR}{P + R},
\end{align*}
where $P$ is boundary precision, $R$ is boundary recall and $F$ is the combined F-measure. As we are interested in accurate predictions, predicted boundary pixels should be at most 1 pixel away from ground-truth boundary pixels to be correct.

\subsection{Evaluation on Natural Video Sequences Involving Agent-based Motion.}
\myparagraph{Dataset and Training.}
We use the VSB100 dataset which contains 101 videos with a maximum 121 frames each.  The training set consists of 40 videos and the test set consists of 60 videos. The videos contain a wide range of objects of different sizes and shapes, including vehicles, humans and animals. The videos also have a wide variety of both object and camera motion. We use the hierarchical video segmentation algorithm in \cite{khoreva2016improved} to segment these videos. The output is a ultra-metric contour map (ucm). Boundaries higher in the hierarchy typically correspond to semantically coherent entities like animals, vehicles etc and therefore their motion corresponds to object/camera motion.  We discard boundaries belonging to the lowest level of the hierarchy (corresponding to an over-segmentation), as they are temporally very unstable. We use the ucm hierarchy as a confidence measure on boundary location at a pixel.

\myparagraph{Experimental Settings and Baselines.} \label{par:mb}
The models are trained to predict boundaries of segmented VSB100 videos. Recall that, the ground-truth boundaries (ucm) in VSB100 have different confidence values. Thus, we threshold the predictions before comparison to the groundtruth. We vary the threshold to obtain a precision-recall curve and report the area under the curve (AUC) along with the best F-measure across all thresholds. We include a ``Last Input'' baseline by using the last input frame as constant prediction and a ``Optical flow'' baseline. As many boundaries do not change between frames in the videos of VSB100, the last input is a strong baseline especially when we are predicting one step into the future. In case of the optic flow baseline, the optic flow is calculated between the last two input frames (at $t$ - 1 and $t$) using the Epic flow method of \cite{revaud2015epicflow}. The boundary pixels at time $t$ are propagated using the calculated flow to generate predictions at $t$ + 1 to $t$ + 8.

\myparagraph{Results on VSB100.}
We perform an ablation study of our CMSC model and we compare to, \begin{enumerate*} \item A convolutional single scale model (CSS) \item A convolutional multi-scale model (CMS) \end{enumerate*}, in addition to the baselines. Both models do not have a context. We report the quantitative results in Figure \ref{fig:aucvsb100} and Figure \ref{fig:bestfvsb100} and the qualitative results in Figure \ref{fig:cframes}. \\
{\textit{Quantitative evaluation:} } In the short term the CMS model (green lines) performs well. However, our CMSC (red lines) performs best in the longer term (both having the same number of parameters). This demonstrates the importance of the context for long-term prediction. The good performance of both of the mutli-scale models (CMS and CMSC) versus the single scale CSS model, shows that multiple scales lead to more accurate predictions. The performance advantage of our CMSC model over the last input baseline shows that the model learns to predict boundaries of moving objects while keeping static boundaries intact. The recall of the CMSC model declines with time as the future becomes increasingly uncertain. The poor performance of the ``Optic flow'' baseline is due to inaccurate flow information at object boundaries.   \\
{\textit{Qualitative evaluation:} } The boundaries produced by our CMSC model are sharp whenever the motion is smooth, e.g. the predictions in Figure \ref{fig:cframes}. However, the models are not able to deal with high uncertainty in the long-term often due to non-deterministic motion. The models in such situations react by blurring the boundaries, as a consequence of using the L2 training loss. While predicting recursively, this leads to loss of boundary confidence and eventual vanishing boundaries. The ``Optic flow'' baseline produces discontinuous (jagged) boundaries. (See Appendix for more examples). Next we evaluate and compare RGB prediction to boundary prediction.

\myparagraph{RGB verses Boundary Prediction.}
We report the sharpness of RGB frames (of VSB100) predicted by the adversarial model of \cite{mathieu2015deep} (fine-tuned on VSB100) using the Laplacian measure \cite{krotkov2012active} in Figure \ref{fig:shapness}. The Laplacian measure pools the gradient information of the image. We observe that the model of \cite{mathieu2015deep} makes increasingly blurry predictions into the future. We also compare the mean squared error of RGB predictions of \cite{mathieu2015deep} and predicted boundaries of our CMSC model in Figure \ref{fig:mse}. We see a sharper increase in the error of RGB predictions compared to boundaries in the long term.

  \begin{table}[h]
  \setlength{\tabcolsep}{3pt}
  \centering
  \begin{tabular}{@{}ccccc@{}}\\
    \toprule
    \textbf{Step}  &\textbf{Last Input} &\textbf{CMS} &\textbf{CMSC-BL} &\textbf{CMSC}  \\
    \hline 
    $t$ + 1      &0.141 &0.282 & 0.957  &\textbf{0.987} \\
    $t$ + 5      &0.038 &0.101 & 0.841  &\textbf{0.900} \\
    $t$ + 20     &0.002 &0.066 & 0.347  &\textbf{0.632} \\
    \bottomrule
  \end{tabular}
  \caption{Evaluation on single ball billiard table worlds.}
  \label{bf_one}
\end{table}

 \begin{table*}[t]
  \centering
  \begin{tabular}{@{}cccccccccc@{}}\\
\toprule
    &\multicolumn{3}{c}{ -- {\bf Evaluation on two ball worlds} --} & \multicolumn{3}{c}{ -- {\bf Evaluation on three ball worlds} -- } & \multicolumn{3}{c}{ -- {\bf Evaluation on six ball worlds} -- }                \\
     \textbf{Step}  & \textbf{Last Input} & \textbf{CMSC-1B}  & \textbf{CMSC} & \textbf{Last Input} & \textbf{CMSC-2B}  & \textbf{CMSC} & \textbf{Last Input} & \textbf{CMSC-3B}  & \textbf{CMSC}  \\
    \hline 
    $t$ + 1      &0.246 &0.966 &\textbf{0.969} &0.246  &0.967 &\textbf{0.968} &0.250  &0.962 &\textbf{0.964} \\
    $t$ + 5      &0.114 &0.848 &\textbf{0.896} &0.118  &0.890 &\textbf{0.892} &0.130  &\textbf{0.875} &0.866\\
    $t$ + 20     &0.101 &0.612 &\textbf{0.681} &0.090  &0.664 &\textbf{0.700} &0.115  &0.511 &\textbf{0.600}\\
    \bottomrule
  \end{tabular}
    \caption{Evaluation on complex billiard table worlds.}
  \label{bf_mult}
\end{table*}

\begin{figure*}[t]
    \centering
  \begin{tabular}{ C{1.8cm}C{1.8cm}C{1.8cm}C{1.8cm}C{1.8cm}C{1.8cm}C{1.8cm}C{1.8cm} }
        \includegraphics[width=0.10\textwidth]{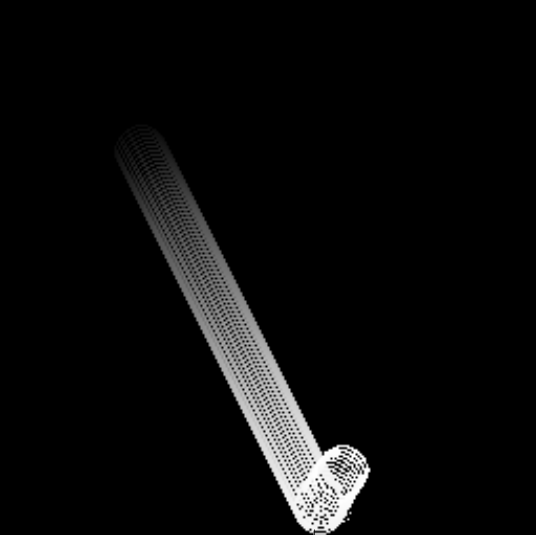} &   
        \includegraphics[width=0.10\textwidth]{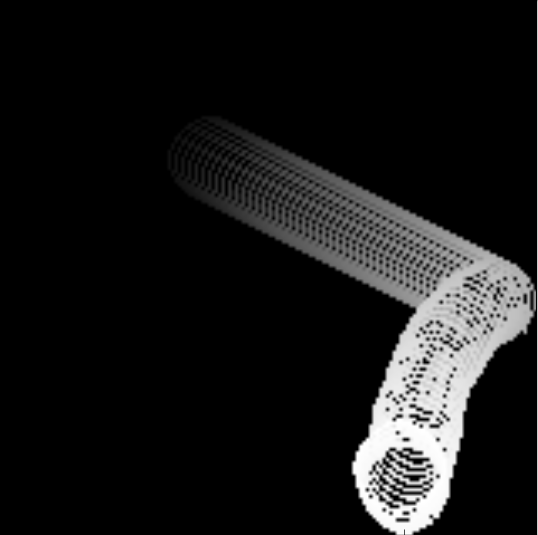} &
        \includegraphics[width=0.10\textwidth]{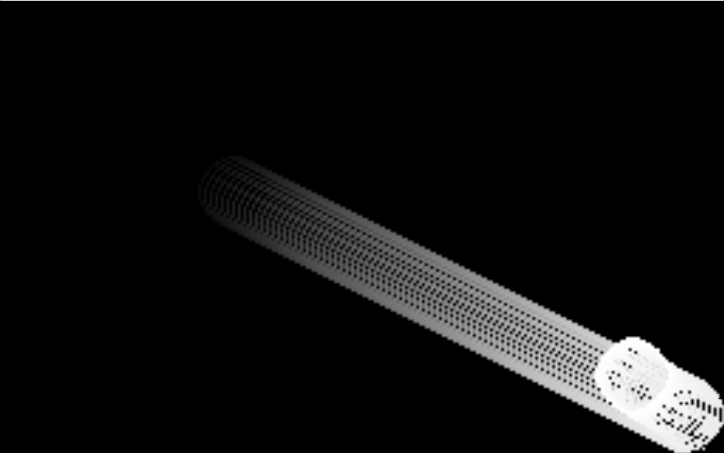} &
        \includegraphics[width=0.10\textwidth]{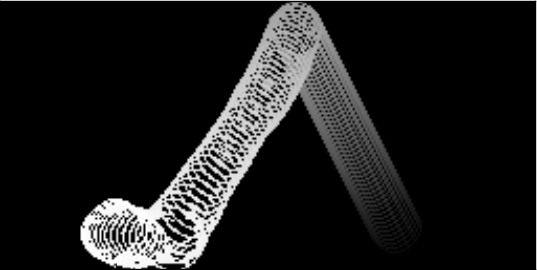} &
        \includegraphics[width=0.10\textwidth]{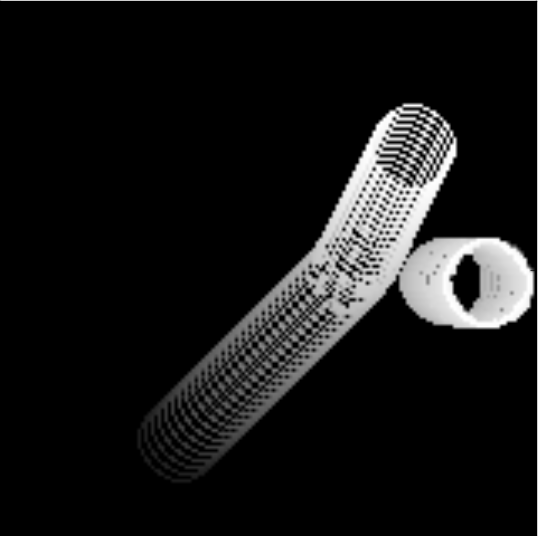} &   
        \includegraphics[width=0.10\textwidth]{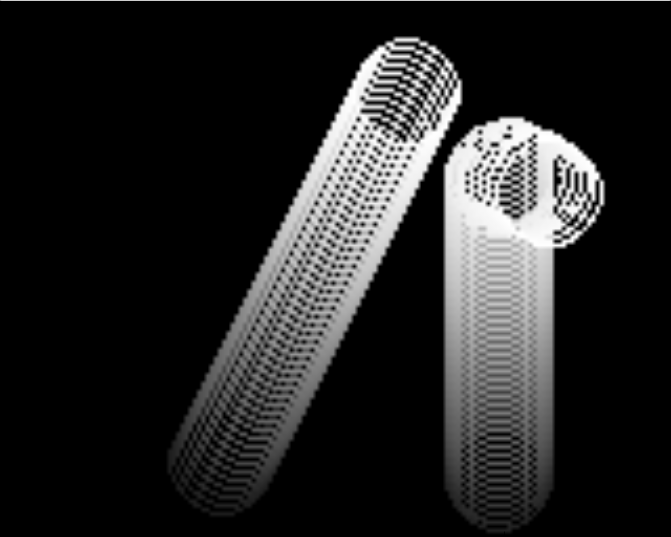} &   
        \includegraphics[width=0.10\textwidth]{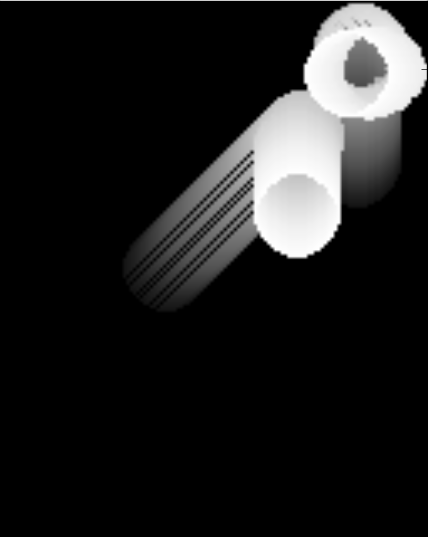} &   
        \includegraphics[width=0.10\textwidth]{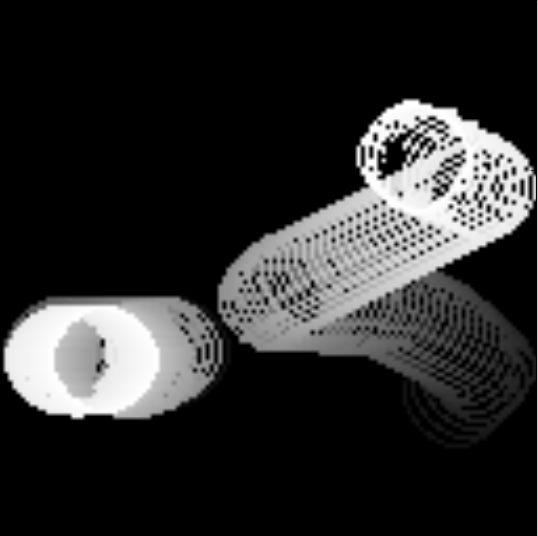} \\ 
        \end{tabular}
    \caption{Trails produced by super-imposing predicted boundaries on synthetic sequences.}
    \label{fig:trails}
\end{figure*}
\begin{figure*}[h!]
  
  \centering
  \begin{tabular}{ C{3.5cm}C{3.5cm}C{3.5cm}C{3.5cm} }
    
    \includegraphics[width=0.18\textwidth]{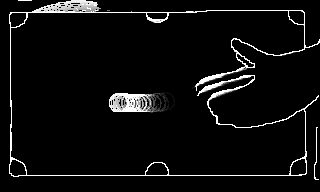} &
    \includegraphics[width=0.18\textwidth]{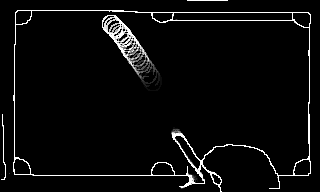} &
    \includegraphics[width=0.18\textwidth]{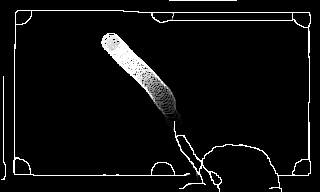}&
    \includegraphics[width=0.18\textwidth]{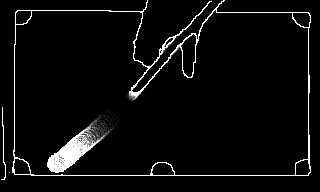} \\
    
    Trail up to $t$ + 20 & Trail up to $t$ + 20 &  Trail up to $t$ + 50 &  Trail up to $t$ + 50 \\
    
    \end{tabular}
  \caption{Trails produced by super-imposing predicted boundaries on real sequences. }
  \label{fig:bframes}
\end{figure*}

\subsection{Evaluation on Physics-based Motion.}
Motion in the videos in the VSB100 dataset is frequently very complex  as agent's actions quickly become non-deterministic and hence increasingly uncertain. Therefore, we also look at physics-based motion, which is still challenging yet it factors out the aforementioned issues. In this scenario, we evaluate the long-term prediction performance of the models on real and synthetic billiard ball sequences. We begin by describing our dataset.

\myparagraph{Synthetic Data Generation.}
The synthetic billiard ball sequences are sampled from worlds which consists of balls moving on a frictionless surface with a border, akin to a billiard table. We used pygame to create such worlds and sample boundary images from them. The output images contain  boundaries that can stem from ball(s) or the table and have binary confidence measure (indicating a boundary at that location). During evaluation, as the target is always a binary image, we report only the best F-measure obtained by thresholding the predicting boundary images and varying the threshold parameter. 
\label{sssec:data_gen_syn}
We sampled synthetic billiard sequences using the following parameters.
\begin{enumerate*}
    \item {\it Table size}: Side length randomly sampled from \{96,128,160,192,256\} pixels.
    \item {\it Ball velocity}: Randomly sampled from [\{-3,..,3\},\{-3,..,3\}] pixels.
    \item {\it Ball size}: Constant, with a radius of 13 pixels.
    \item {\it Initial Position}: Uniformly over the table surface.
\end{enumerate*}

\myparagraph{Real Data Collection.}
We captured a novel data-set of real billiard sequences on a mini-billiard table. Frame rate was set to 120 per second to minimize motion blur. Each sequence consists of an actor (not visible) striking the ball with a cue stick once. The motion in the sequences of the dataset are that of the cue stick and the balls. We produce boundary images using the method of \cite{Man+16a}.

\myparagraph{Evaluation on synthetic single ball worlds.}
We generate a training set using parameters mentioned previously. However, to keep our training set as diverse as possible we prefer short sequences. We restrict each sequence to a maximum length of one or two collisions with walls and set a 50\% bias of the initial position of the balls being 40 pixels from the walls. We sample 500 such sequences and train our models on these sequences. We then test the models on 30 independent test sequences. We again include the ``last input'' baseline as a constant predictor . We also include a ``blind'' Convolutional multi-scale Context model (CMSC-BL), which cannot see the table borders. This is a strong baseline as starting from 42\% frames in the test set, there are no ball-wall collisions 20 steps into the future. To beat this baseline, our models need to learn the physics of ball-wall collisions.  We report the results in Table \ref{bf_one}.

Our CMSC model performs the best with accurate predictions 20 time-steps into the future -- also exceeding the ``blind'' version (CMSC-BL) that cannot handle ball-wall collisions. The model without a context CMS, produces inaccurate results at patch borders and thus suffers heavily especially at larger time-steps.

\myparagraph{Evaluation on synthetic two and three ball worlds.}
Worlds with more than one ball also involve harder to model physics of ball-ball collisions. To evaluate the models on such worlds we sample 100 training sequences each with two, three and six balls respectively with a maximum length of 200 frames. We use a curriculum learning approach \cite{bengio2009curriculum}, where we initialize the models with the weights learned on single, two and three ball worlds respectively.  We test the models on 30 independent sequences containing two, three and six balls respectively. We report the results in Table \ref{bf_mult}. In each case, we also include CMSC models trained on single ball worlds (CMSC-1B), two ball worlds (CMSC-2B) and three ball worlds (CMSC-3B) respectively as baselines. To beat these strong baselines learning the physics of ball-ball collisions is necessary as in case of our two-ball and three-ball test sets, there are no ball-ball and 3-ball collisions 20 steps into the future for 92\% and 98\% of the starting frames (and no 6-ball collisions). Again, we see accurate prediction by the CMSC model even at 20 time-steps in the future.

\myparagraph{Prediction over Very Long Time Scales.}
Although we evaluate only 20 timesteps into the future in Table \ref{bf_one} and Table \ref{bf_mult}, our models are stable over longer time-horizons. In Figure \ref{fig:trails}, we predict 100 timesteps and visualize the boundary images by trails obtained by superposition. We notice a few failure cases where a ball reverse direction mid table and the ball(s) get deformed or disappear. 

\begin{table}[h]
\vspace{-0.5cm}
  \centering
  \resizebox{\linewidth}{!}{
    \begin{tabular}{@{}ccccc@{}}\\
  \toprule
    \textbf{Step}  &\textbf{Last Input} &\textbf{CMSC} &\textbf{Last Input(M)} &\textbf{CMSC(M)}\\
    \hline
    $t$ + 1      &0.890 &0.850 &0.126 &0.570 \\ 
    $t$ + 5      &0.855 &0.804 &0.085 &0.541 \\ 
    $t$ + 20     &0.844 &0.746 &0.087 &0.497 \\ 
    \bottomrule
  \end{tabular}
  }
   \caption{Evaluation on real billiard sequences (M-masked).}
  \label{bf_real}
\end{table}

 \begin{table*}[t]
  \centering
     \resizebox{\textwidth}{!}{
  \begin{tabular}{@{}cccccccccc@{}}\\
\toprule
    &\multicolumn{3}{c}{-------- {\bf PSNR} --------} & \multicolumn{3}{c}{-------- {\bf Sharpness Loss} --------} & \multicolumn{3}{c}{-------- {\bf Laplacian Measure} --------}                \\
    \midrule
     \textbf{Step}  & \textbf{RGB prediction} & \textbf{De-blurring} & \textbf{Fusion (Ours)} & \textbf{RGB prediction} & \textbf{De-blurring} & \textbf{Fusion (Ours)} & \textbf{RGB prediction} & \textbf{De-blurring} & \textbf{Fusion (Ours)}    \\
     \midrule
     \multicolumn{10}{c}{\bf VSB100} \\
        \midrule 
        $t$ + 2      &24.4 &24.5 &\textbf{25.1} &18.5 &18.5 &\textbf{18.6} &0.142 &0.139 &\textbf{0.155}  \\
        $t$ + 3      &22.2 &22.9 &\textbf{23.1} &18.2 &18.2 &\textbf{18.3} &0.121 &0.109 &\textbf{0.127} \\
        $t$ + 4      &20.4 &21.7 &\textbf{22.3} &18.1 &18.1 &\textbf{18.2} &0.103 &0.114 &\textbf{0.118} \\
        \midrule
        
    \multicolumn{10}{c}{\bf UCF101} \\
        \midrule 
        $t$ + 2      &26.5 &27.7 &\textbf{28.2} &21.4 &21.5 &\textbf{21.7} &0.101 &0.122 &\textbf{0.136} \\
        $t$ + 3      &23.4 &25.1 &\textbf{25.2} &20.5 &20.8 &\textbf{20.9} &0.095 &0.093 &\textbf{0.102} \\
        $t$ + 4      &21.4 &23.4 &\textbf{23.8} &20.4 &20.5 &\textbf{20.6} &0.089 &0.101 &\textbf{0.112} \\
        \bottomrule
  \end{tabular}}
    \caption{Evaluation of our Fusion scheme. PSNR, Sharpness Loss and Laplacian measure: Higher is better. }
  \label{tab:fusion}
\end{table*}

\begin{figure*}[t]
  
  \centering
  \renewcommand{\arraystretch}{0.2}
  \begin{tabular}{ C{0.8cm}C{1.58cm}C{1.58cm}C{1.58cm}C{1.58cm}C{1.58cm}C{1.58cm}C{1.58cm}C{1.58cm} }
  \toprule
  \scriptsize{\textbf{Timestep}} & \scriptsize{\textbf{RGB prediction}} & \scriptsize{\textbf{De-blurring}} &  \scriptsize{\textbf{Fusion (Ours)}} & \scriptsize{\textbf{Groundtruth}} & \scriptsize{\textbf{RGB prediction} } & \scriptsize{\textbf{De-blurring}} &  \scriptsize{\textbf{Fusion (Ours)}} & \scriptsize{\textbf{Groundtruth}} \\
  \midrule
   t + 2 &
    \includegraphics[width=0.110\textwidth, height=0.055\textheight]{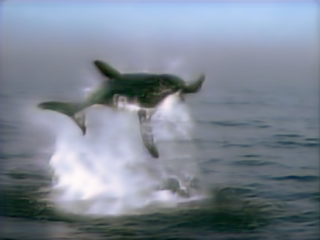} &
    \includegraphics[width=0.110\textwidth, height=0.055\textheight]{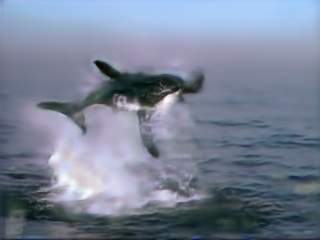} &
    \includegraphics[width=0.110\textwidth, height=0.055\textheight]{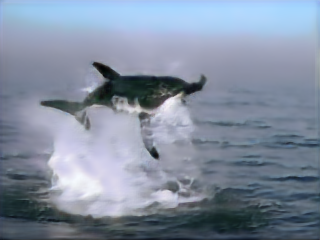} &
    \includegraphics[width=0.110\textwidth, height=0.055\textheight]{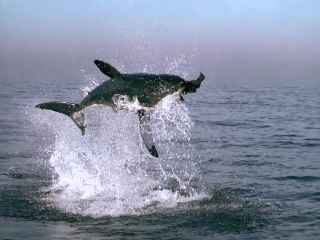} &
    \includegraphics[width=0.110\textwidth, height=0.055\textheight]{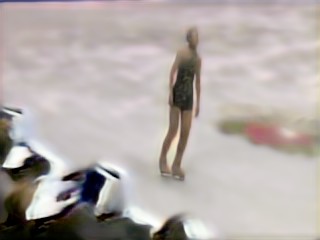}&
    \includegraphics[width=0.110\textwidth, height=0.055\textheight]{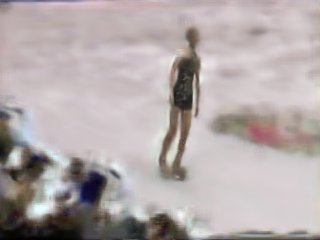}&
    \includegraphics[width=0.110\textwidth, height=0.055\textheight]{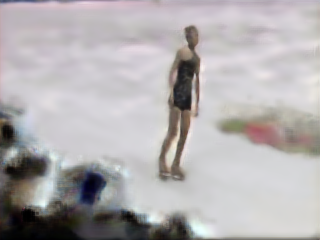}&
    \includegraphics[width=0.110\textwidth, height=0.055\textheight]{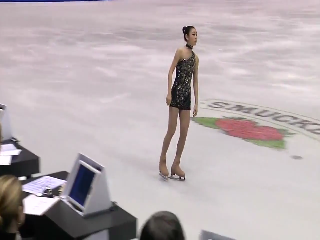}\\
    
    t + 4 &
    \includegraphics[width=0.110\textwidth, height=0.055\textheight]{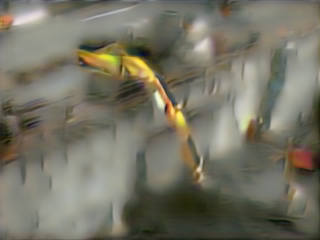} &
    \includegraphics[width=0.110\textwidth, height=0.055\textheight]{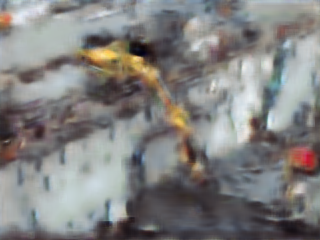} &
    \includegraphics[width=0.110\textwidth, height=0.055\textheight]{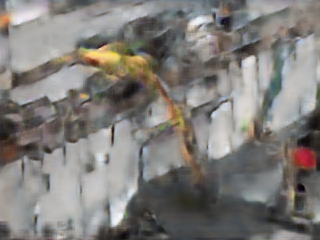} &
    \includegraphics[width=0.110\textwidth, height=0.055\textheight]{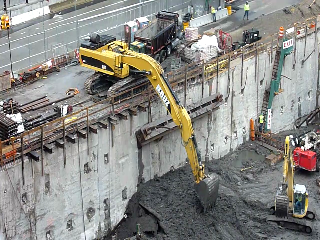} &
    \includegraphics[width=0.110\textwidth, height=0.055\textheight]{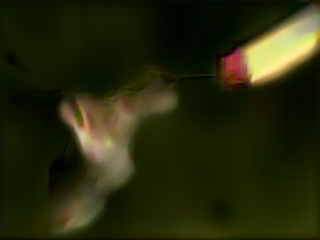}&
    \includegraphics[width=0.110\textwidth, height=0.055\textheight]{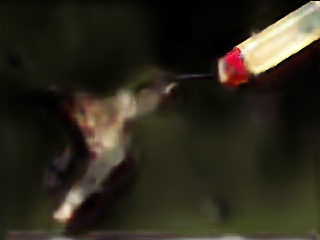}&
    \includegraphics[width=0.110\textwidth, height=0.055\textheight]{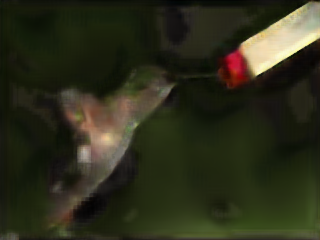}&
    \includegraphics[width=0.110\textwidth, height=0.055\textheight]{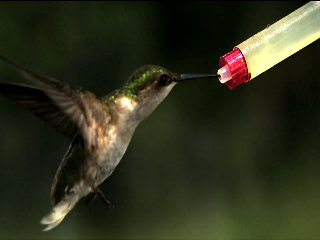}\\
    \midrule
    
    t + 2 &
    \includegraphics[width=0.110\textwidth, height=0.055\textheight]{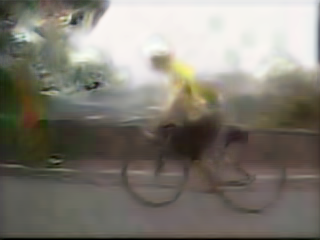} &
    \includegraphics[width=0.110\textwidth, height=0.055\textheight]{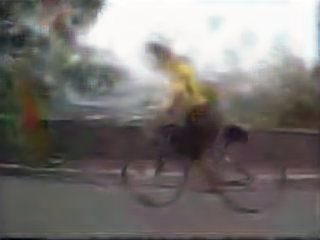} &
    \includegraphics[width=0.110\textwidth, height=0.055\textheight]{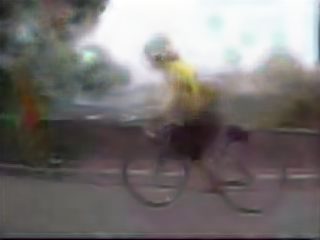} &
    \includegraphics[width=0.110\textwidth, height=0.055\textheight]{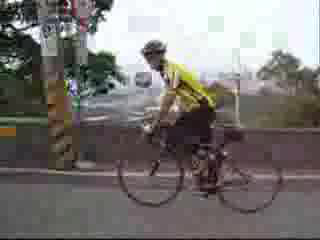} &
    \includegraphics[width=0.110\textwidth, height=0.055\textheight]{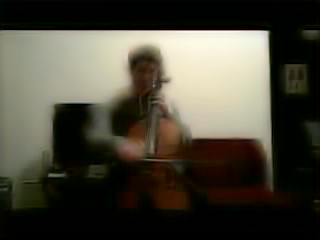}&
    \includegraphics[width=0.110\textwidth, height=0.055\textheight]{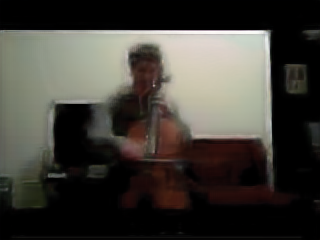}&
    \includegraphics[width=0.110\textwidth, height=0.055\textheight]{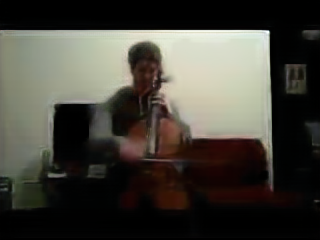}&
    \includegraphics[width=0.110\textwidth, height=0.055\textheight]{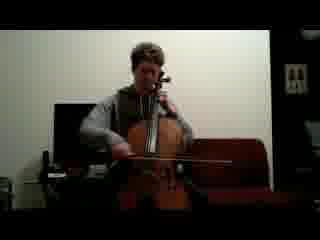}\\
    
    t + 4 &
    \includegraphics[width=0.110\textwidth, height=0.055\textheight]{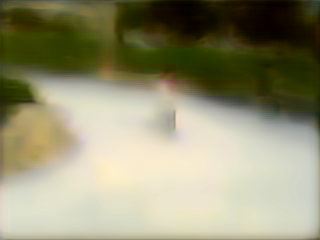} &
    \includegraphics[width=0.110\textwidth, height=0.055\textheight]{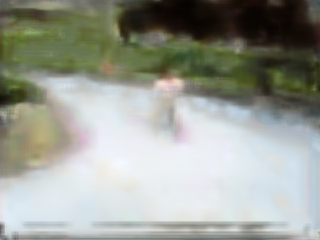} &
    \includegraphics[width=0.110\textwidth, height=0.055\textheight]{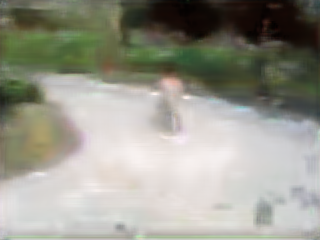} &
    \includegraphics[width=0.110\textwidth, height=0.055\textheight]{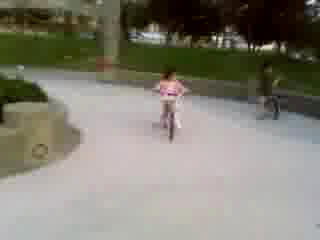} &
    \includegraphics[width=0.110\textwidth, height=0.055\textheight]{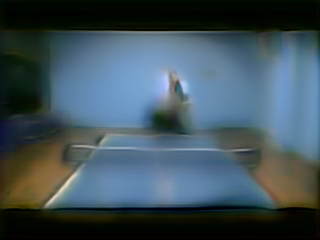}&
    \includegraphics[width=0.110\textwidth, height=0.055\textheight]{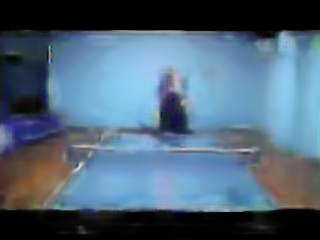}&
    \includegraphics[width=0.110\textwidth, height=0.055\textheight]{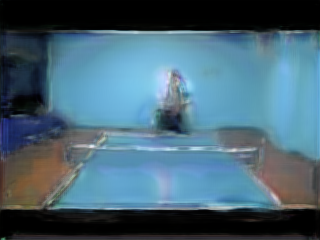}&
    \includegraphics[width=0.110\textwidth, height=0.055\textheight]{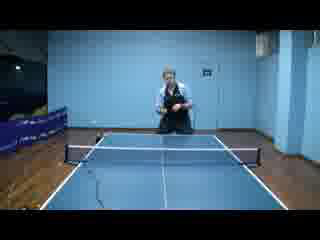}\\
    
    \bottomrule
    
    \end{tabular}
  \caption{Sharpening RGB predictions using our Fusion scheme on VSB100 (top two rows) and on UCF101 (bottom two rows).}
  \label{fig:rgbframes}
\end{figure*}

\begin{figure}[h]
    \centering
        \includegraphics[width=\linewidth]{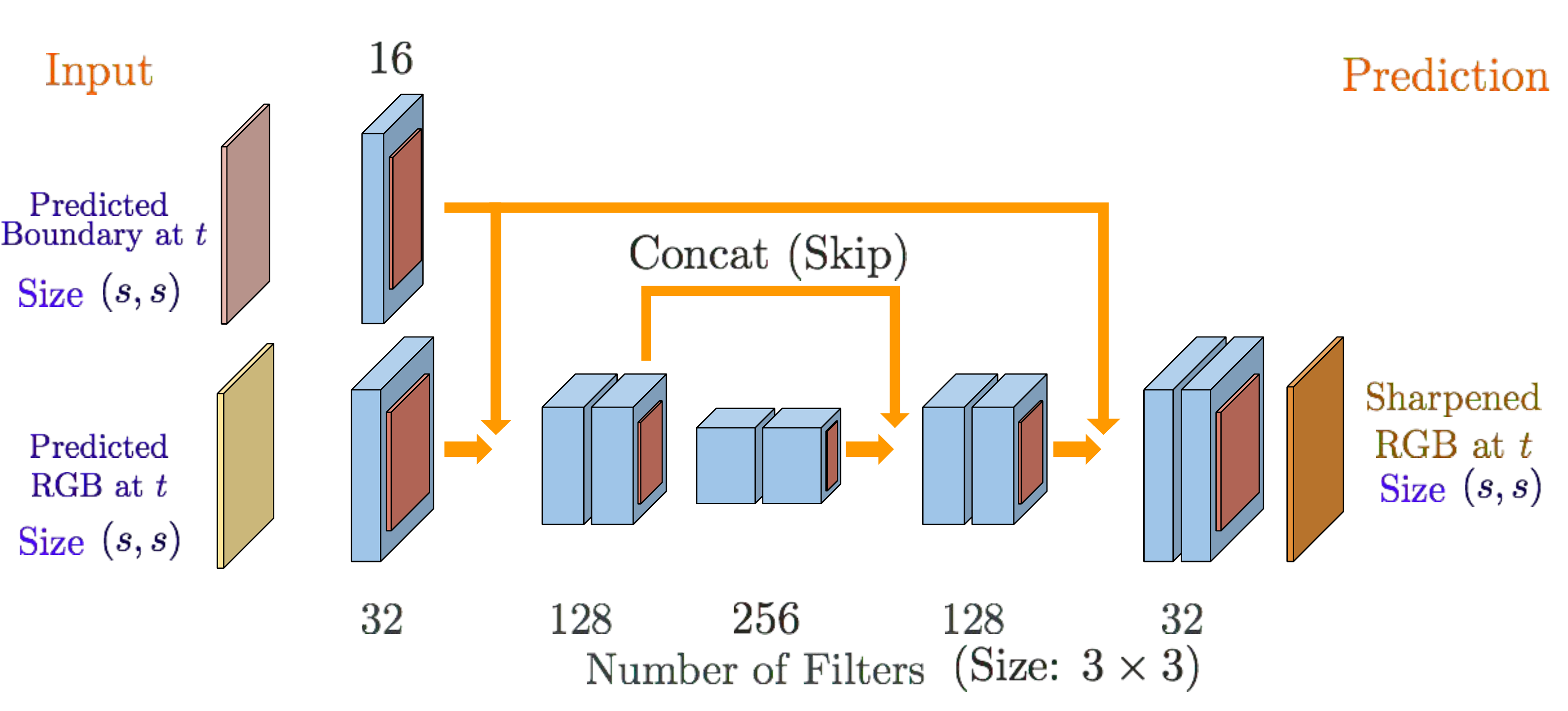}    
        \caption{Our fusion model architecture.}
    \label{fig:fusionarch}
\end{figure}

\myparagraph{Evaluation on Real Billiard Sequences.}
Prediction on real billiard table sequences is a challenging test for our models. The table fabric causes rapid deceleration of the ball (compared to the constant velocity in the synthetic sequences). Spin is sometimes inadvertently introduced and a segmentation algorithm applied on the observed frames introduces artifacts. The boundaries are not always consistent across frames of a sequence and they are jagged and change shape.  We collect 350 real billiard sequences, with one ball, as our training set. To deal with deceleration, we experiment with increasing the number of input frames. We train our CMSC model with six input frames and pre-train on our synthetic one ball training set. We report the results of evaluation (F-measure as before) on 30 independent sequences in Table \ref{bf_real}. As many boundaries (e.g table borders) remain static the last input baseline performs very well. For fair comparison we use a mask obtained with a ball tracker,  Our method is able to propagate the motion of the ball and beats the last input baseline in the masked case. We show qualitative results in Table \ref{fig:bframes} as trails, where our model predicts 20 and 50 time-steps into the future.

\section{Sharpening RGB Predictions with Fusion}
The sharp boundaries produced by our models raise the prospect of sharpening RGB predictions in a fusion scheme. We present our fusion architecture in Figure \ref{fig:fusionarch}, which fuses RGB predictions of \cite{mathieu2015deep} with our boundaries. Note that, our approach can be used on top of any RGB frame prediction method and unlike \cite{villegas2017learning} is video domain agnostic. It is inspired by prior work \cite{eigen2013restoring,mao2016image} on deblurring/denoising. Like these models our fusion model is fully convolutional. Resolution is maintained by skip connections, as in \cite{mao2016image}.  Our fusion model takes as input the predicted RGB and boundaries at each timestep and is trained with L2 loss.

\myparagraph{Datasets and metrics.} We evaluate on both VSB100 and UCF101 datasets. We randomly select 30 and 20 videos from VSB100 to train our CMSC model and our fusion model. We test on the remaining 50 videos. Similarly we randomly select 1000, 500 (training) and 1000 (test) videos from UCF101. The UCF101 train/test set was segmented using the method of \cite{Man+16a}. We use PSNR, the sharpness loss measure from \cite{mathieu2015deep} and the Laplacian measure as evaluation metrics.

\myparagraph{Baselines.} We include a baseline de-blurring model. It has the same architecture as our fusion model, except for the top block. This baseline aims to de-blur RGB predictions without observing our predicted boundaries.

\myparagraph{Evaluation.} We observe improved and sharper RGB predictions (see Table \ref{tab:fusion}) \footnote{ Corresponding results in Table 5 in \cite{mathieu2015deep}. We do not use motion masking as we would like our model to keep still boundaries intact. }. Our fusion model learns to reintroduce lost high frequency information.

\FloatBarrier
\section{Conclusion}
We propose the novel task of boundary prediction and demonstrate accurate results with our CMSC model. We argue for the key design choices,  
\begin{enumerate*}
    \item A wide receptive field allowing the model to learn complex spatio-temporal dependencies.
    \item Accurate prediction at each time-step with a fully convolutional setup without any bottleneck layers.
    \item The context which allows for information sharing thus leading to global consistency. 
\end{enumerate*}
 We obtain sharp predictions using L2 loss (in contrast to RGB prediction, which leads to very blurry results with L2 loss). Predictions by our CMSC model on diverse scenarios shows that it developed a data-driven model of future boundary motions over long time horizons. This includes dynamics of moving agents and  billiard balls. Moreover, while not being our primary goal, our predicted boundaries lead to sharper RGB video predictions via a fusion-based approach.


\spacing{0.97}{
\bibliography{egbib}

\begin{thebibliography}{}

\bibitem[\protect\citeauthoryear{Arbelaez \bgroup et al\mbox.\egroup
  }{2011}]{amfm_pami2011}
Arbelaez, P.; Maire, M.; Fowlkes, C.; and Malik, J.
\newblock 2011.
\newblock Contour detection and hierarchical image segmentation.
\newblock {\em TPAMI}.

\bibitem[\protect\citeauthoryear{Baillargeon}{1994}]{baillargeon1994infants}
Baillargeon, R.
\newblock 1994.
\newblock How do infants learn about the physical world?
\newblock {\em Current Directions in Psychological Science}.

\bibitem[\protect\citeauthoryear{Baillargeon}{2004}]{baillargeon2004infants}
Baillargeon, R.
\newblock 2004.
\newblock Infants' physical world.
\newblock {\em Current directions in psychological science} 13(3):89--94.

\bibitem[\protect\citeauthoryear{Battaglia \bgroup et al\mbox.\egroup
  }{2016}]{battaglia2016interaction}
Battaglia, P.; Pascanu, R.; Lai, M.; Rezende, D.~J.; et~al.
\newblock 2016.
\newblock Interaction networks for learning about objects, relations and
  physics.
\newblock In {\em Advances in Neural Information Processing Systems},
  4502--4510.

\bibitem[\protect\citeauthoryear{Bengio \bgroup et al\mbox.\egroup
  }{2009}]{bengio2009curriculum}
Bengio, Y.; Louradour, J.; Collobert, R.; and Weston, J.
\newblock 2009.
\newblock Curriculum learning.
\newblock In {\em ICML}.

\bibitem[\protect\citeauthoryear{Chang, Wei, and Fisher}{2013}]{chang2013video}
Chang, J.; Wei, D.; and Fisher, J.
\newblock 2013.
\newblock A video representation using temporal superpixels.
\newblock In {\em CVPR}.

\bibitem[\protect\citeauthoryear{Denton \bgroup et al\mbox.\egroup
  }{2015}]{denton2015deep}
Denton, E.~L.; Chintala, S.; Szlam, A.; and Fergus, R.
\newblock 2015.
\newblock Deep generative image models using a laplacian pyramid of adversarial
  networks.
\newblock In {\em NIPS}.

\bibitem[\protect\citeauthoryear{Eigen, Krishnan, and
  Fergus}{2013}]{eigen2013restoring}
Eigen, D.; Krishnan, D.; and Fergus, R.
\newblock 2013.
\newblock Restoring an image taken through a window covered with dirt or rain.
\newblock In {\em CVPR}.

\bibitem[\protect\citeauthoryear{Finn, Goodfellow, and
  Levine}{2016}]{finn2016unsupervised}
Finn, C.; Goodfellow, I.; and Levine, S.
\newblock 2016.
\newblock Unsupervised learning for physical interaction through video
  prediction.
\newblock In {\em NIPS},  64--72.

\bibitem[\protect\citeauthoryear{Fragkiadaki \bgroup et al\mbox.\egroup
  }{2016}]{fragkiadaki2015learning}
Fragkiadaki, K.; Agrawal, P.; Levine, S.; and Malik, J.
\newblock 2016.
\newblock Learning visual predictive models of physics for playing billiards.
\newblock {\em ICLR}.

\bibitem[\protect\citeauthoryear{Galasso \bgroup et al\mbox.\egroup
  }{2013}]{galasso2013unified}
Galasso, F.; Nagaraja, N.; Cardenas, T.; Brox, T.; and Schiele, B.
\newblock 2013.
\newblock A unified video segmentation benchmark: Annotation, metrics and
  analysis.
\newblock In {\em CVPR}.

\bibitem[\protect\citeauthoryear{Galasso \bgroup et al\mbox.\egroup
  }{2014}]{galasso2014spectral}
Galasso, F.; Keuper, M.; Brox, T.; and Schiele, B.
\newblock 2014.
\newblock Spectral graph reduction for efficient image and streaming video
  segmentation.
\newblock In {\em CVPR}.

\bibitem[\protect\citeauthoryear{Jain \bgroup et al\mbox.\egroup
  }{2007}]{jain2007supervised}
Jain, V.; Murray, J.~F.; Roth, F.; Turaga, S.; Zhigulin, V.; Briggman, K.~L.;
  Helmstaedter, M.~N.; Denk, W.; and Seung, H.~S.
\newblock 2007.
\newblock Supervised learning of image restoration with convolutional networks.
\newblock In {\em CVPR}.

\bibitem[\protect\citeauthoryear{Kalchbrenner \bgroup et al\mbox.\egroup
  }{2017}]{kalchbrenner2016video}
Kalchbrenner, N.; Oord, A. v.~d.; Simonyan, K.; Danihelka, I.; Vinyals, O.;
  Graves, A.; and Kavukcuoglu, K.
\newblock 2017.
\newblock Video pixel networks.
\newblock {\em ICML}.

\bibitem[\protect\citeauthoryear{Khoreva \bgroup et al\mbox.\egroup
  }{2016}]{khoreva2016improved}
Khoreva, A.; Benenson, R.; Galasso, F.; Hein, M.; and Schiele, B.
\newblock 2016.
\newblock Improved image boundaries for better video segmentation.
\newblock In {\em ECCV Workshop}.

\bibitem[\protect\citeauthoryear{Krotkov}{2012}]{krotkov2012active}
Krotkov, E.~P.
\newblock 2012.
\newblock {\em Active computer vision by cooperative focus and stereo}.
\newblock Springer Science \& Business Media.

\bibitem[\protect\citeauthoryear{Lerer, Gross, and
  Fergus}{2016}]{lerer2016learning}
Lerer, A.; Gross, S.; and Fergus, R.
\newblock 2016.
\newblock Learning physical intuition of block towers by example.
\newblock In {\em ICML}.

\bibitem[\protect\citeauthoryear{Li, Leonardis, and Fritz}{2017}]{li2016fall}
Li, W.; Leonardis, A.; and Fritz, M.
\newblock 2017.
\newblock Visual stability prediction for robotic manipulation.
\newblock In {\em IEEE International Conference on Robotics and Automation
  (ICRA)}.
\newblock to appear.

\bibitem[\protect\citeauthoryear{Liang \bgroup et al\mbox.\egroup
  }{2017}]{liang2017dual}
Liang, X.; Lee, L.; Dai, W.; and Xing, E.~P.
\newblock 2017.
\newblock Dual motion gan for future-flow embedded video prediction.
\newblock {\em ICCV}.

\bibitem[\protect\citeauthoryear{Liu \bgroup et al\mbox.\egroup
  }{2017}]{liu2017video}
Liu, Z.; Yeh, R.; Tang, X.; Liu, Y.; and Agarwala, A.
\newblock 2017.
\newblock Video frame synthesis using deep voxel flow.
\newblock {\em ICCV}.

\bibitem[\protect\citeauthoryear{Maninis \bgroup et al\mbox.\egroup
  }{2016}]{Man+16a}
Maninis, K.; Pont-Tuset, J.; Arbel\'{a}ez, P.; and Gool, L.~V.
\newblock 2016.
\newblock Convolutional oriented boundaries.
\newblock In {\em ECCV}.

\bibitem[\protect\citeauthoryear{Mao, Shen, and Yang}{2016}]{mao2016image}
Mao, X.-J.; Shen, C.; and Yang, Y.-B.
\newblock 2016.
\newblock Image restoration using convolutional auto-encoders with symmetric
  skip connections.
\newblock {\em arXiv preprint arXiv:1606.08921}.

\bibitem[\protect\citeauthoryear{Mathieu, Couprie, and
  LeCun}{2016}]{mathieu2015deep}
Mathieu, M.; Couprie, C.; and LeCun, Y.
\newblock 2016.
\newblock Deep multi-scale video prediction beyond mean square error.
\newblock {\em ICLR}.

\bibitem[\protect\citeauthoryear{Michalski, Memisevic, and
  Konda}{2014}]{michalski2014modeling}
Michalski, V.; Memisevic, R.; and Konda, K.
\newblock 2014.
\newblock Modeling deep temporal dependencies with recurrent grammar cells.
\newblock In {\em NIPS}.

\bibitem[\protect\citeauthoryear{Ochs, Malik, and
  Brox}{2014}]{ochs2014segmentation}
Ochs, P.; Malik, J.; and Brox, T.
\newblock 2014.
\newblock Segmentation of moving objects by long term video analysis.
\newblock {\em TPAMI}.

\bibitem[\protect\citeauthoryear{Patraucean, Handa, and
  Cipolla}{2015}]{patraucean2015spatio}
Patraucean, V.; Handa, A.; and Cipolla, R.
\newblock 2015.
\newblock Spatio-temporal video autoencoder with differentiable memory.
\newblock {\em arXiv preprint arXiv:1511.06309}.

\bibitem[\protect\citeauthoryear{Ranzato \bgroup et al\mbox.\egroup
  }{2014}]{ranzato2014video}
Ranzato, M.; Szlam, A.; Bruna, J.; Mathieu, M.; Collobert, R.; and Chopra, S.
\newblock 2014.
\newblock Video (language) modeling: a baseline for generative models of
  natural videos.
\newblock {\em arXiv:1412.6604}.

\bibitem[\protect\citeauthoryear{Revaud \bgroup et al\mbox.\egroup
  }{2015}]{revaud2015epicflow}
Revaud, J.; Weinzaepfel, P.; Harchaoui, Z.; and Schmid, C.
\newblock 2015.
\newblock Epicflow: Edge-preserving interpolation of correspondences for
  optical flow.
\newblock In {\em CVPR}.

\bibitem[\protect\citeauthoryear{Srivastava, Mansimov, and
  Salakhutdinov}{2015}]{srivastava2015unsupervised}
Srivastava, N.; Mansimov, E.; and Salakhutdinov, R.
\newblock 2015.
\newblock Unsupervised learning of video representations using lstms.
\newblock In {\em ICML}.

\bibitem[\protect\citeauthoryear{Sutskever, Hinton, and
  Taylor}{2009}]{sutskever2009recurrent}
Sutskever, I.; Hinton, G.~E.; and Taylor, G.~W.
\newblock 2009.
\newblock The recurrent temporal restricted boltzmann machine.
\newblock In {\em NIPS}.

\bibitem[\protect\citeauthoryear{Villegas \bgroup et al\mbox.\egroup
  }{2017}]{villegas2017learning}
Villegas, R.; Yang, J.; Zou, Y.; Sohn, S.; Lin, X.; and Lee, H.
\newblock 2017.
\newblock Learning to generate long-term future via hierarchical prediction.
\newblock {\em ICML, 2017}.

\bibitem[\protect\citeauthoryear{Watters \bgroup et al\mbox.\egroup
  }{2017}]{watters2017visual}
Watters, N.; Tacchetti, A.; Weber, T.; Pascanu, R.; Battaglia, P.; and Zoran,
  D.
\newblock 2017.
\newblock Visual interaction networks.
\newblock {\em arXiv preprint arXiv:1706.01433}.

\bibitem[\protect\citeauthoryear{Wertheimer}{1923}]{wertheimer1923laws}
Wertheimer, M.
\newblock 1923.
\newblock Laws of organization in perceptual forms.
\newblock {\em A source book of Gestalt Psychology}.

\end{thebibliography}
\bibliographystyle{aaai}
}

\spacing{1.00}
\newpage

\section{Appendix}
We include here additional details of our model and results.

\subsection{Further Details of the CMSC Model.} We include the internal details of each level $\text{L}_{k}$ of our CMSC model in Table \ref{tab:cmsc_details1}. We include the details of the type of layer, type specific details including the number and size of convolutional filters and pooling/upsampling layers, the non-linearity (activation) used after every layer, input and output of each layer.

\begin{table}[h]
\centering
\resizebox{\linewidth}{!}{
\begin{tabular}{ccccccc}
\toprule
Layer & Type & Filters & Size & Activation & Input & Output \\
\midrule
$\text{In}_1$ & Input & & & & & C1 \\
\midrule
$\text{C}_1$ & Conv & 32 & 3$\times$3 & \emph{ReLU} & $\text{In}_1$ & $\text{C}_2$ \\
$\text{C}_2$ & Conv & 32 & 3$\times$3 & \emph{ReLU} & $\text{C}_1$ & $\text{P}_1$ \\
$\text{P}_1$ & MaxPool & & 2$\times$2 & & $\text{C}_2$ & $\text{C}_3$ \\
\midrule
$\text{C}_3$ & Conv & 64 & 3$\times$3 & \emph{ReLU} & $\text{P}_1$ & $\text{C}_2$ \\
$\text{C}_4$ & Conv & 64 & 3$\times$3 & \emph{ReLU} & $\text{C}_3$ & $\text{P}_2$ \\
$\text{P}_2$ & MaxPool & & 2$\times$2 & & $\text{C}_4$ & $\text{C}_5$ \\
\midrule
$\text{C}_5$ & Conv & 128 & 3$\times$3 & \emph{ReLU} & $\text{P}_2$ & $\text{C}_6$ \\
$\text{C}_6$ & Conv & 128 & 3$\times$3 & \emph{ReLU} & $\text{C}_5$ & $\text{U}_1$ \\
$\text{U}_1$ & UpSample & & 2$\times$2 & & $\text{C}_6$ & $\text{C}_7$ \\
\midrule
$\text{C}_7$ & Conv & 64 & 3$\times$3 & \emph{ReLU} & $\text{U}_1$ & $\text{C}_8$ \\
$\text{C}_8$ & Conv & 64 & 3$\times$3 & \emph{ReLU} & $\text{C}_7$ & $\text{U}_2$ \\
$\text{U}_2$ & UpSample & & 2$\times$2 & & $\text{C}_8$ & $\text{C}_9$ \\
\midrule
$\text{C}_9$ & Conv & 32 & 3$\times$3 & \emph{ReLU} & $\text{U}_2$ & $\text{C}_{10}$ \\
$\text{C}_{10}$ & Conv & 1 & 3$\times$3 & \emph{tanh} & $\text{C}_9$ &  \\
\bottomrule
\end{tabular}
}
\caption{Internal details of each level $\text{L}_{k}$ of our CMSC model. Conv stands for 2D convolution, MaxPool stands for 2D max pooling and UpSample stands for 2D upsampling operations.}
\label{tab:cmsc_details1}
\end{table}

We include details of each of the four levels of our CMSC model in Table \ref{tab:cmsc_details2}. We include details of the scale (resolution) at which each level operates and the input to each level. We use the same notation as in the main paper, $\text{X}_{k}$ denotes the input boundary image at scale $k\times k$ and $\hat{\text{O}}(\text{L}_{k})$ is the upsampled (by factor 2$\times$2) output of level $\text{L}_{k}$. The final output is produced by the $\text{L}_{96}$ level.

\begin{table}[h]
\centering
\begin{tabular}{cccc}
\toprule
Level & Scale & Input \\
\midrule
$\text{L}_{12}$ & 12$\times$12 & $\text{X}_{12}$ \\
\midrule
$\text{L}_{24}$ & 24$\times$24 & $\left\{ \text{X}_{24}, \, \hat{\text{O}}(\text{L}_{12}) \right\}$ \\
\midrule
$\text{L}_{48}$ & 48$\times$48 & $\left\{ \text{X}_{48}, \, \hat{\text{O}}(\text{L}_{24}) \right\}$ \\
\midrule
$\text{L}_{96}$ & 96$\times$96 & $\left\{ \text{X}_{96}, \, \hat{\text{O}}(\text{L}_{48}) \right\}$ \\
\bottomrule
\end{tabular}
\caption{Details of the levels in our CMSC model.}
\label{tab:cmsc_details2}
\end{table}

The central 32$\times$32 patch of the output, produced by the $\text{L}_{96}$ level, is considered valid and used to generate the boundary image at the next time-step.

\subsection{Results on Moving MNIST.}

\begin{figure*}[!h]
  
  \centering
  \renewcommand{\arraystretch}{0.2}
  \begin{tabular}{ C{0.95cm}C{0.95cm}C{0.95cm}C{0.95cm}C{0.95cm}C{0.95cm}C{0.95cm}C{0.95cm}C{0.95cm}C{0.95cm}C{0.95cm}C{0.95cm}C{0.95cm}C{0.95cm}C{0.95cm}C{0.95cm} }

    \includegraphics[width=0.070\textwidth]{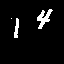} &
    \includegraphics[width=0.070\textwidth]{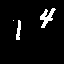} &
    \includegraphics[width=0.070\textwidth]{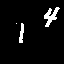} &
    \includegraphics[width=0.070\textwidth]{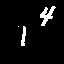} &
    \includegraphics[width=0.070\textwidth]{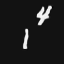}&
    \includegraphics[width=0.070\textwidth]{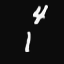}&
    \includegraphics[width=0.070\textwidth]{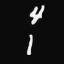}&
    \includegraphics[width=0.070\textwidth]{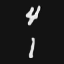}&
    \includegraphics[width=0.070\textwidth]{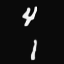}&
    \includegraphics[width=0.070\textwidth]{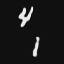}&
    \includegraphics[width=0.070\textwidth]{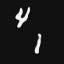}&
    \includegraphics[width=0.070\textwidth]{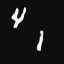}\\
    
    \includegraphics[width=0.070\textwidth]{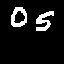} &
    \includegraphics[width=0.070\textwidth]{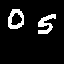} &
    \includegraphics[width=0.070\textwidth]{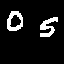} &
    \includegraphics[width=0.070\textwidth]{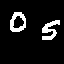} &
    \includegraphics[width=0.070\textwidth]{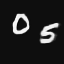}&
    \includegraphics[width=0.070\textwidth]{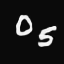}&
    \includegraphics[width=0.070\textwidth]{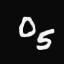}&
    \includegraphics[width=0.070\textwidth]{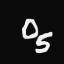}&
    \includegraphics[width=0.070\textwidth]{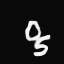}&
    \includegraphics[width=0.070\textwidth]{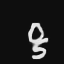}&
    \includegraphics[width=0.070\textwidth]{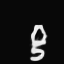}&
    \includegraphics[width=0.070\textwidth]{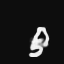}\\
    
    \includegraphics[width=0.070\textwidth]{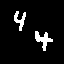} &
    \includegraphics[width=0.070\textwidth]{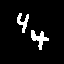} &
    \includegraphics[width=0.070\textwidth]{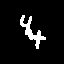} &
    \includegraphics[width=0.070\textwidth]{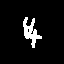} &
    \includegraphics[width=0.070\textwidth]{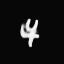}&
    \includegraphics[width=0.070\textwidth]{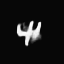}&
    \includegraphics[width=0.070\textwidth]{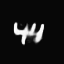}&
    \includegraphics[width=0.070\textwidth]{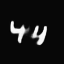}&
    \includegraphics[width=0.070\textwidth]{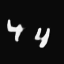}&
    \includegraphics[width=0.070\textwidth]{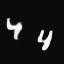}&
    \includegraphics[width=0.070\textwidth]{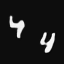}&
    \includegraphics[width=0.070\textwidth]{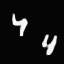}\\
    
    \includegraphics[width=0.070\textwidth]{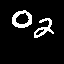} &
    \includegraphics[width=0.070\textwidth]{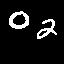} &
    \includegraphics[width=0.070\textwidth]{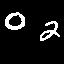} &
    \includegraphics[width=0.070\textwidth]{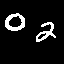} &
    \includegraphics[width=0.070\textwidth]{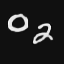}&
    \includegraphics[width=0.070\textwidth]{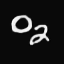}&
    \includegraphics[width=0.070\textwidth]{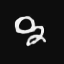}&
    \includegraphics[width=0.070\textwidth]{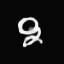}&
    \includegraphics[width=0.070\textwidth]{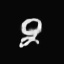}&
    \includegraphics[width=0.070\textwidth]{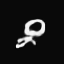}&
    \includegraphics[width=0.070\textwidth]{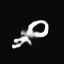}&
    \includegraphics[width=0.070\textwidth]{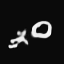}\\
    
    \addlinespace[+1ex]
    \multicolumn{4}{c}{$\xleftarrow{\makebox[1.50cm]{}}$ Observation $\xrightarrow{\makebox[1.50cm]{}}$} &\multicolumn{8}{c}{$\xleftarrow{\makebox[4.35cm]{}}$ Prediction $\xrightarrow{\makebox[4.37cm]{}}$} \\
    
    \addlinespace[+1ex]
    {\textbf{$t-3$}} &{\textbf{$t-2$}} &{\textbf{$t-1$}} &{\textbf{$t$}} 
    &{\textbf{$t+1$}} &{\textbf{$t+2$}}  &{\textbf{$t+3$}}  &{\textbf{$t+4$}}
    &{\textbf{$t+5$}} &{\textbf{$t+6$}}  &{\textbf{$t+7$}} &{\textbf{$t+8$}}  \\
    \addlinespace[+1ex]
    
    \end{tabular}
  \caption{Example predictions on the moving MNIST dataset by our CMS model.}
  \label{fig:mnist}
\end{figure*}

We include results on the moving MNIST dataset \cite{srivastava2015unsupervised} to help compare our Convolutional Multi-Scale architecture against other frame prediction architectures. This dataset is suitable for this task because like boundary images the images in the moving MNIST dataset are also in the same domain [0,1]. However, we do not refer to this dataset in the main article as moving MNIST digits behave very differently compared to object boundaries. As the moving MNIST digits are of fixed size of 64x64 pixels, we do not use a context and thus use our CMS model for evaluation (which observes the full input image). This allows fair comparison against \cite{srivastava2015unsupervised,patraucean2015spatio,kalchbrenner2016video} which do not have a context and observes the full input image. We report quantitative results for prediction one time-step into the future (as in \cite{srivastava2015unsupervised,patraucean2015spatio}) in Table \ref{tab:mnist} using the Cross Entropy Loss \cite{srivastava2015unsupervised}. Our CMS model outperforms \cite{srivastava2015unsupervised,patraucean2015spatio}. Moreover, qualitative results in Figure \ref{fig:mnist} shows that our CMS model predicts accurately eight time-steps into the future. The highly complex model from \cite{kalchbrenner2016video} performs better. However, this comparison shows that our CMS model compares favorably against other frame prediction models, while beating models with comparable number of parameters.

\begin{table}[h]
\vspace{-0.5cm}
  \centering
   \resizebox{\linewidth}{!}{
    \begin{tabular}{@{}cc@{}}\\
  \toprule
    \textbf{Model}  &\textbf{Cross Entropy Loss} \\
    \hline
    \cite{srivastava2015unsupervised}      & 341.2 \\ 
    \cite{patraucean2015spatio}      &179.8 \\
    Our CMS & \textbf{165.0}  \\
    \cite{kalchbrenner2016video} & 87.6\\
    \bottomrule
  \end{tabular}
  }

   \caption{Evaluation on moving MNIST}
  \label{tab:mnist}
\end{table}

\begin{figure*}[t]
  
  \centering
  \begin{tabular}{ C{3.4cm}C{3.4cm}C{3.4cm}C{3.4cm} }
  
   \toprule
   \textbf{Last Observation: $t$} & \textbf{Prediction: $t$ + 1} & \textbf{Prediction: $t$ + 2} &  \textbf{Prediction: $t$ + 4} \\
    \midrule
    \multicolumn{4}{c}{\textbf{``Optic Flow'' baseline.}} \\
    \midrule
    \includegraphics[width=0.21\textwidth]{"images/fig4/1_gt"} &
    \includegraphics[width=0.21\textwidth]{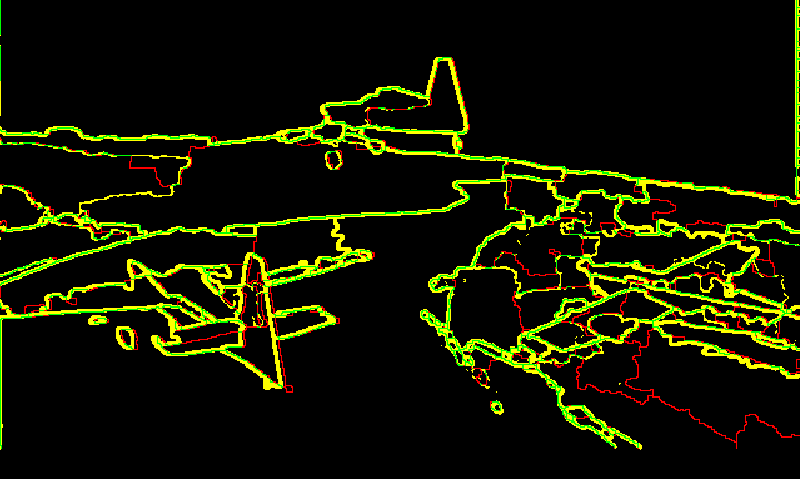} &
    \includegraphics[width=0.21\textwidth]{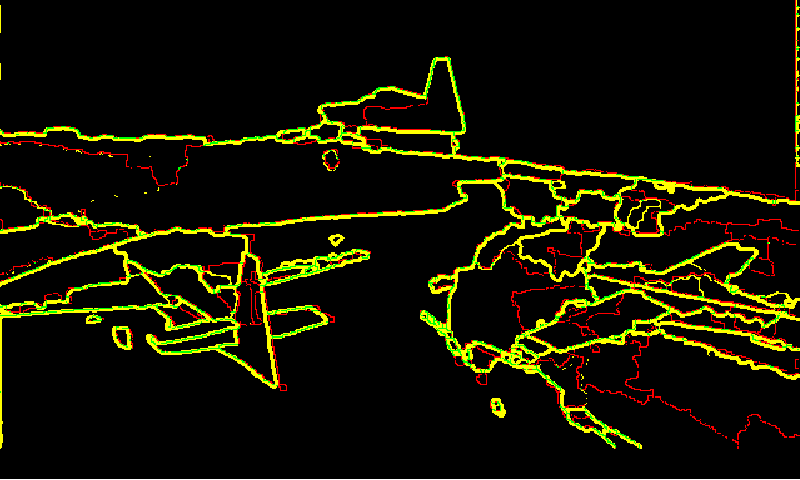} &
    \includegraphics[width=0.21\textwidth]{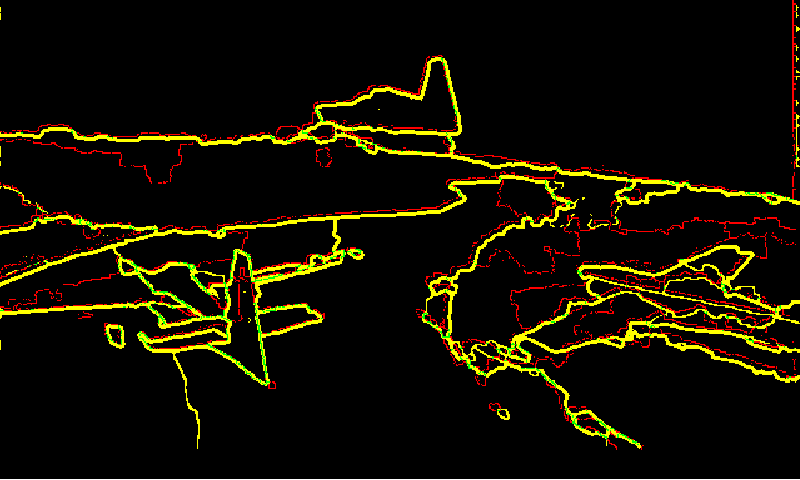} \\
    \midrule
     \multicolumn{4}{c}{\textbf{Convolutional Single Scale (CSS).}} \\
     \midrule
    \includegraphics[width=0.21\textwidth]{"images/fig4/1_gt"}&
    \includegraphics[width=0.21\textwidth]{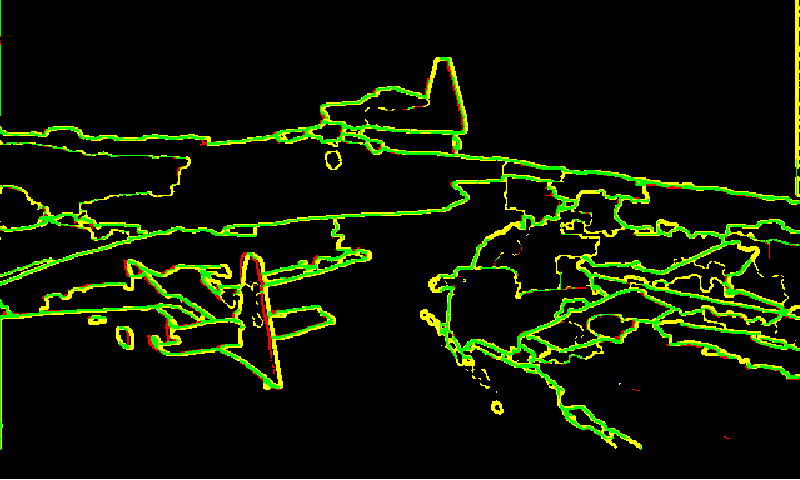}&
    \includegraphics[width=0.21\textwidth]{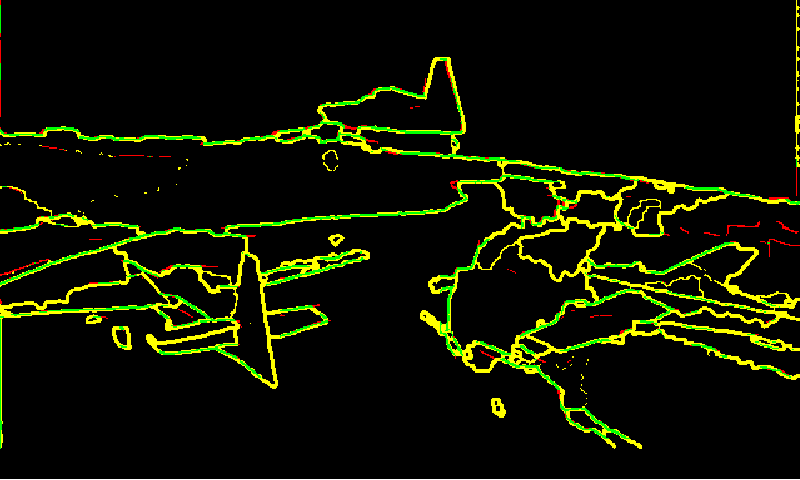}&
    \includegraphics[width=0.21\textwidth]{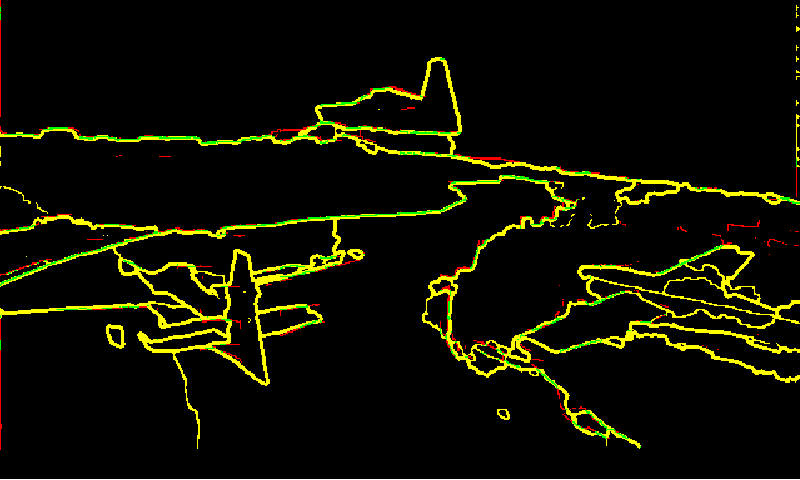}\\
    \midrule
    \multicolumn{4}{c}{\textbf{Convolutional Multi Scale (CMS).}} \\
    \midrule
    \includegraphics[width=0.21\textwidth]{"images/fig4/1_gt"}&
    \includegraphics[width=0.21\textwidth]{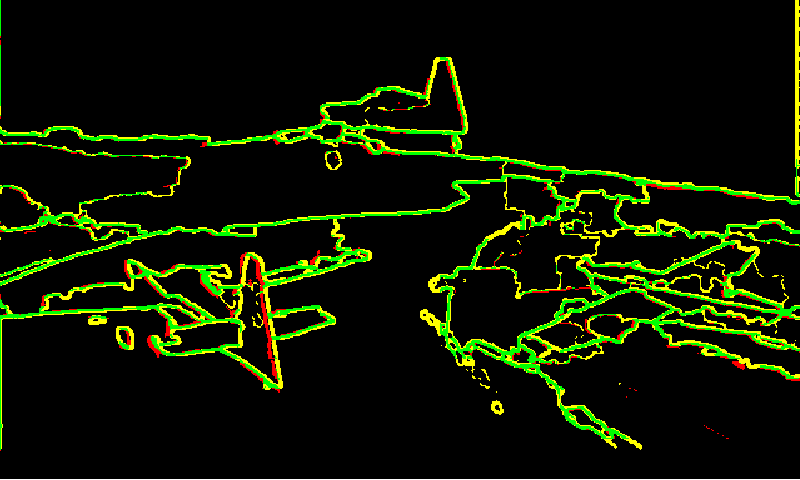}&
    \includegraphics[width=0.21\textwidth]{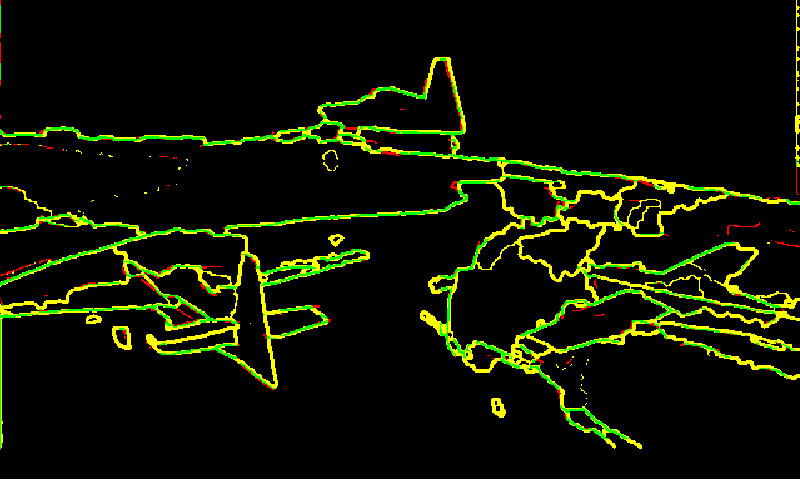}&
    \includegraphics[width=0.21\textwidth]{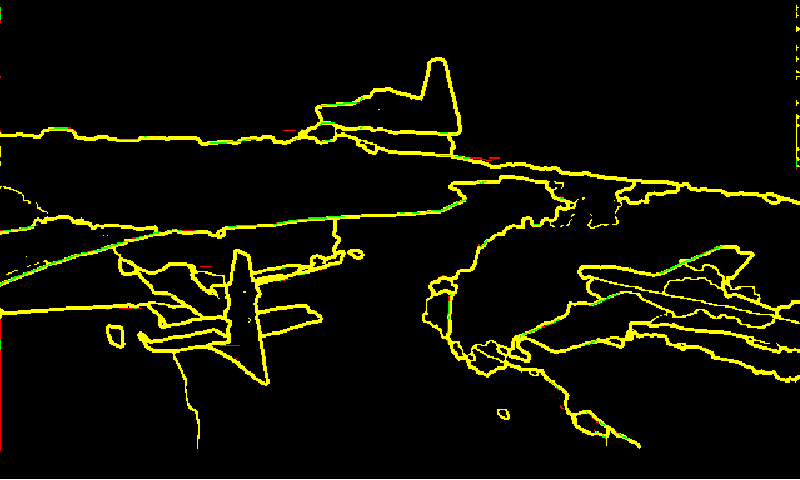}\\
    \midrule
    \multicolumn{4}{c}{\textbf{Convolutional Multi Scale Context (CMSC).}} \\
    \midrule
    \includegraphics[width=0.21\textwidth]{"images/fig4/1_gt"}&
    \includegraphics[width=0.21\textwidth]{"images/fig4/1_t_1"}&
    \includegraphics[width=0.21\textwidth]{"images/fig4/1_t_2"}&
    \includegraphics[width=0.21\textwidth]{"images/fig4/1_t_4"}\\
    \bottomrule
    \end{tabular}
  \caption{Prediction on \emph{airplane} sequence from VSB100. Correct boundaries predictions are encoded in green. Missed boundaries are encoded in yellow. Wrong boundaries are encoded in red.}
  \label{fig:exex}
\end{figure*}

\subsection{Additional Results on VSB100.}
Here, we show predictions at one, two and four steps ($t$ + 1, $t$ + 2, $t$ + 4) in the future from a fixed time point on the airplane sequences of VSB100. We show the predictions in Figure \ref{fig:exex}. We use the same color coding as in the main article. That is, correct boundaries predictions are encoded in green, missed boundaries are encoded in yellow and wrong boundaries are encoded in red.

As expected from the quantitative performance in Figure 5 of the main article, the ``Optic flow'' baseline does not perform well. This method incorrectly translates the boundaries which lead to many boundaries being missed especially at t + 4. Compared to our CMSC model, the CSS and CMS models are unable to propagate motion in the long-term, leading to the disappearance of boundaries at t + 4. This highlights the importance of the context. 

\myparagraph{Running Time.} Running time is GPU model and video resolution dependent. On the Nvidia Titan X GPU, our CMSC model takes approximately 16 hours to train on the VSB100 and real billiards datasets and 10 hours on synthetic billiards (1 ball) dataset. During the test phase, prediction of one future frame of VSB100 (640$\times$480 pixels) takes 1.03 seconds, synthetic billiards (256$\times$256) 136 milliseconds and real billiards (320$\times$240) 168 milliseconds on average.

\end{document}